\documentclass[10pt,journal,compsoc]{IEEEtran}
\ifCLASSOPTIONcompsoc
  \usepackage[nocompress]{cite}
\else
  \usepackage{cite}
\fi
\ifCLASSINFOpdf
\else
\fi

\usepackage[colorlinks=false]{hyperref}
\usepackage[cmex10]{amsmath}
\usepackage{mdwmath}
\usepackage{mdwtab}
\usepackage{amsthm}
\usepackage{amssymb}
\usepackage{amsmath}
\usepackage{graphicx}
\usepackage{subfig}
\usepackage{array}
\usepackage{multirow}
\usepackage{booktabs}
\usepackage{url}
\usepackage{algorithm}
\usepackage{algpseudocode}
\usepackage{algpascal}
\usepackage{epstopdf}
\usepackage{color}
\usepackage{bm}

\begin{document}
%
\title{Multi-view Deep One-class Classification: \\ A Systematic Exploration}


\author{Siqi Wang, Jiyuan Liu, Guang Yu, Xinwang Liu, Sihang Zhou, En Zhu, Yuexiang Yang, Jianping Yin
\IEEEcompsocitemizethanks{
\IEEEcompsocthanksitem S.~Wang, J.~Liu, G.~Yu, X.~Liu, S.~Zhou, E.~Zhu and Y. Yang are with College of Computer, National University of Defense Technology (NUDT), Changsha, 410073, China. E-mail: \{wangsiqi10c, liujiyuan13, xinwangliu, enzhu, yyx\}@nudt.edu.cn, \{yuguangnudt, sihangjoe\}@gmail.com.
\IEEEcompsocthanksitem J.~Yin is with Dongguan University of Technology, Dongguan, 523808, China. E-mail: jpyin@dgut.edu.cn.
\IEEEcompsocthanksitem S.~Wang and J.~Liu contributed equally to this work.
}
\thanks{Manuscript received April 26, 2021.}}

%


\IEEEtitleabstractindextext{%
\begin{abstract}
One-class classification (OCC), which models one single positive class and distinguishes it from the negative class, has been a long-standing topic with pivotal application to realms like anomaly detection. As modern society often deals with massive high-dimensional complex data spawned by multiple sources, it is natural to consider OCC from the perspective of multi-view deep learning. However, it has not been discussed by the literature and remains an unexplored topic. Motivated by this blank, this paper makes four-fold contributions: First, to our best knowledge, this is the first work that formally identifies and formulates the multi-view deep OCC problem. Second, we take recent advances in relevant areas into account and systematically devise eleven different baseline solutions for multi-view deep OCC, which lays the foundation for research on multi-view deep OCC. Third, to remedy the problem that limited benchmark datasets are available for multi-view deep OCC, we extensively collect existing public data and process them into more than 30 new multi-view benchmark datasets via multiple means, so as to provide a publicly available evaluation platform for multi-view deep OCC.
Finally, by comprehensively evaluating the devised solutions on benchmark datasets, we conduct a thorough analysis on the effectiveness of the designed baselines, and hopefully provide other researchers with beneficial guidance and insight to multi-view deep OCC. Our data and codes are opened at \url{https://github.com/liujiyuan13/MvDOCC-datasets} and \url{https://github.com/liujiyuan13/MvDOCC-code} respectively to facilitate future research.

\end{abstract}

}

\maketitle

\IEEEdisplaynontitleabstractindextext

%
\IEEEpeerreviewmaketitle

\section{Introduction}\label{sec:introduction}

\IEEEPARstart{O}ne-class classification (OCC) \cite{moya1993one,Tax2001One} is a classic machine learning problem that only data from one single class (termed as the positive class) are given during training, while very few or zero data from other classes (collectively called as the negative class) are available. 
For inference, the trained OCC model is expected to be capable of distinguishing whether an incoming datum is from the given positive class or the unknown negative class. 
In literature, OCC is also referred as anomaly detection \cite{ruff2018deep}, novelty detection \cite{scikit-learn} or out-of-distribution detection \cite{sedlmeier2020policy} and etc.
It catches the eyes of researchers from both academia and industry for its pervasive applications in practice.
For instance, a public video surveillance system has easy access to data of normal daily events, whilst abnormal events like robbery or vehicle intrusion are rare and extremely hard to predict. 
Therefore, it is often unrealistic to collect sufficient anomalous data from the negative class, which constitutes to one typical application scenario for OCC. 
Besides, OCC techniques have also been widely applied in various realms like information retrieval \cite{manevitz2001one}, fault detection \cite{shin2005one}, authorship verification \cite{koppel2004authorship}, enhanced multi-class classification \cite{krawczyk2015usefulness} and etc. 

Compared with supervised binary/multi-class classification, OCC remains a special and challenging problem. This is mainly due to the lack of training data from the negative class, which makes it impossible to directly train a classifier by discriminating the positive and the negative class. 
Therefore, OCC often resorts to some unsupervised learning methods. 
Meanwhile, OCC is also different from classic unsupervised learning tasks like clustering, as OCC handles training data that share a common positive label, which serves as weak supervision. 
So far, various solutions have been proposed \cite{khan2014one} to tackle OCC and will be reviewed in Sec. \ref{sec:related_work} and supplementary material.

Nevertheless, as the modern society has witnessed an explosive development in data acquisition capabilities, people find it increasingly difficult to perform learning tasks, which include but are not confined to OCC, with modern data via classic methods. 
In this paper, we will focus on two of the most important challenges: 
First, unlike traditional data, modern data such as images are often endowed with high dimension and complex latent structures. 
classic methods usually fail to exploit such latent information embedded in data, due to their shallow model architectures and limited representation power. 
Second, with significantly enriched sources to acquire data, one object is often described from multiple viewpoints such as different modalities, sensors or angles, which gives birth to a large amount of multi-view data. 
However, classic methods are usually designed for single-view data, and they lack the ability to exploit complementary information and cross-view correlation embedded in multi-view data. 
To handle the above two challenges posed by modern data, an emerging realm named \textit{multi-view deep learning} has attracted surging attention from researchers. 
Specifically, multi-view deep learning resorts to artificial neural networks with deep architecture to conduct layer-wise data abstraction and representation learning \cite{lecun2015deep}, which is proven highly effective in vast applications. 
The remarkable success of deep learning has made it a standard tool to handle massive complex data. 
Meanwhile, multi-view deep learning methods usually leverage multi-view fusion or multi-view alignment techniques \cite{li2018survey} to exploit inter-view information embedded in multi-view data. Multi-view deep learning has already been successfully applied to many tasks \cite{li2018survey,baltruvsaitis2018multimodal,guo2019deep}. 
Therefore, it is quite natural for us to consider the intersection of OCC and multi-view deep learning, i.e. \textit{multi-view deep OCC}. 
Multi-view deep OCC has a large potential to many practical problems, and a straightforward example is the aforementioned video surveillance system that aims to detect anomalies: 
The collected normal events can be described by both RGB and optical flow data, which are both high-dimensional data with rich underlying semantics, while OCC model needs to be trained with such data to build a normality model and discriminate the anomalies.

Although both OCC and multi-view deep learning methods have been thoroughly studied in literature, the problem of multi-view deep OCC has not been formally defined and systematically explored to our best knowledge.
Such a blank constitutes to the biggest motivation of this paper. 
There are three major obstacles when looking into multi-view deep OCC: \textbf{1)} \textit{Above all, the lack of formal formulation of the problem.} Despite its huge application potential in various real-world scenarios, multi-view deep OCC has not been formally identified and formulated, which prevents researchers from giving sufficient attention to this novel but challenging problem. \textbf{2)} \textit{Second, the lack of baseline methods}. Although many attempts have been made to approach multi-view deep learning, they are typically designed for other tasks and therefore not explored for OCC. In the meantime, existing approaches for OCC are merely applicable to the single-view case. \textbf{3)} \textit{Third, the lack of proper benchmark datasets for evaluation}. 
Previous researches usually evaluate OCC models by the ``one \textit{vs.} all'' protocol \cite{golan2018deep}.
For any binary/multi-class benchmark datasets, it assumes data from a certain class as positive, while data from the rest of classes as negative. 
Besides, datasets that are specifically designed for OCC are also proposed recently \cite{Rayana2016}. 
However, frequently-used benchmark datasets in OCC are basically single-view. 
In the meantime, existing multi-view benchmark datasets are often too small, and none of them are specifically designed for the background of OCC. As a result, very few benchmarks are suitable for the evaluation of multi-view deep OCC. 

To bridge the above gaps, this paper, for the first time, formally formulates the problem of multi-view deep OCC, and carries out a systematic study on this new area. Our contributions can be summarized as follows:

\begin{itemize}
    \item To our best knowledge, this is the first work that formally identifies and formulates the multi-view deep OCC problem, which points out a brand new area for both OCC and multi-view deep learning. 
    \item Inspired by recent literature of OCC and multi-view deep learning, we systematically design 11 different solutions to multi-view deep OCC, which provide informative baselines to this problem. We open the implementations of all baselines to facilitate further research on multi-view deep OCC.
    \item To provide a standard evaluation platform of multi-view deep OCC, we extensively collect existing public data and process them into more than 30 new multi-view benchmark datasets via various means. All benchmark datasets are made publicly available.
    \item We comprehensively evaluate the proposed multi-view deep OCC baselines on both constructed and existing multi-view datasets, and conduct in-depth analysis on their performance. 
    It sheds the first light on multi-view deep OCC, and hopefully provides valuable guidance and insights to future research.   
\end{itemize}

\section{Related Work}
\label{sec:related_work}

In this section, we will focus on reviewing deep learning based OCC, multi-view anomaly/outlier and multi-view deep learning, which are the most relevant areas to multi-view deep OCC. In the Sec. 1 and Sec. 2 of supplementary material, we also briefly review classic methods for OCC and multi-view learning due to the limit of space.

\subsection{Deep Learning based OCC}

There is a surging interest to leverage deep neural networks to handle high-dimensional one-class data \cite{pang2020deep}. 
Since only data from a single class are available, the most frequently-used models for deep learning based OCC are generative deep neural networks (DNNs). 
A simple but effective way is to extend the shallow auto-encoder (AE) into a deep one. 
For example, stacked denoising auto-encoder (SDAE) \cite{xu2017detecting} and deep convolutional auto-encoder (DCAE) \cite{hasan2016learning} have been leveraged to perform OCC with raw video data.
Meanwhile, many attempts are also made to improve AE's OCC performance, such as using the ensemble technique \cite{chen2017outlier} and combining AE with energy based model \cite{zhai2016deep}. 
In addition to AE based methods, other popular generative neural networks like generative adversarial networks (GANs) \cite{schlegl2017unsupervised,akcay2018ganomaly} and U-Net \cite{liu2018future,ye2019anopcn} are also actively explored to perform OCC. 
Such generative models typically perform OCC by measuring the reconstruction error of the generated target data, while other methods (e.g. the discriminator outputs and latent representations of GANs) are also explored. 
Apart from generative deep models, several representative discriminative approaches are also proposed recently.
Ruff et al. \cite{ruff2018deep} extend the classic SVDD into deep SVDD (DSVDD), which learns to map latent representations of positive data into a hypersphere with minimal radius. 
Golan et al. \cite{golan2018deep} for the first time leverage self-supervised learning for image OCC. 
They impose multiple geometric transformations to create pseudo classes, which are classified by a discriminative DNN to enable highly effective representation learning. 
Statistics of the discriminative DNN outputs are then used to score each image. 
Bergman et al. \cite{bergman2019classification} further extends self-supervised learning based deep OCC to generic tabular data by introducing random projection for creating pseudo classes. 
Goyal et al. \cite{goyal2020drocc} assume a low-dimensional manifold in given positive data, which can be utilized to sample accurate pseudo outliers to train a discriminative component. 
The detail review can be found in \cite{DBLP:conf/wsdm/PangC021}.
Despite that great progress has been made in deep OCC, existing researches are still limited to the single-view setting.

\subsection{Multi-view Anomaly/Oultier Detection}
Multi-view anomaly/outlier detection is a closely relevant but different area to multi-view deep OCC. The pioneer work of multi-view anomaly/outlier detction is proposed by Gao et al. \cite{gao2011spectral}, while a series of works follow their setup and propose improved solutions \cite{marcos2013clustering,zhao2015dual,iwata2016multi,li2018multi,sheng2019multi}. Although OCC is often discussed in the context of semi-supervised anomaly/outlier detection, multi-view anomaly/outlier detection here is actually a fully unsupervised task. It aims to detect either intra-view outliers (``attribute outlier'') or cross-view inconsistency (``class outlier'') from unlabeled data. 
By contrast, OCC adopts a setup that handles pure data from a single class, which makes them intrinsically different from each other.
Recently, Wang et al. \cite{wang2020towards} for the first time discuss the case of ``semi-supervised multi-view anomaly/detection'', which is essentially multi-view OCC, by a hierarchical Bayesian model. 
Nevertheless, their method cannot perform representation learning. Meanwhile, it is only tested on classic benchmarks and suffers from poor scalability to large data. Thus, their work still has a gap to the multi-view deep OCC discussed in this paper. 



\subsection{Multi-view Deep Learning}

Since classic multi-view learning does not involve representation learning process and lack of the ability to handle with complex modern data, multi-view deep learning has rapidly become an emerging topic. 
Current multi-view deep learning methods are usually categorized into two groups, i.e. \textit{Multi-view fusion} and \textit{multi-view alignment} based methods.
Multi-view fusion based methods fuse the learned representations from different views into a joint representation, which can be realized by either simple operations like max/sum/concatenation \cite{li2018survey}, or sophisticated means like a neural network. 
Specifically, the pioneer work of Ngiam et al. \cite{ngiam2011multimodal} proposes a multi-modal deep auto-encoder for multi-view deep fusion, while Srivastava et al. \cite{srivastava2014multimodal} perform the fusion by multi-modal deep boltzman machine (DBM). 
Such neural network based multi-view fusion can also be conducted on modern neural network architecture like convolutional neural networks (CNNs) \cite{feichtenhofer2016convolutional} and recurrent neural networks (RNNs) \cite{mao2014deep}. 
Latest work from Sun et al. \cite{sun2020multi} employs a multi-view deep Gaussian Process to obtain the joint representation and perform classification. 
Apart from the prevalent neural network based fusion, Zadeh et al. \cite{zadeh2017tensor} propose a novel tensor based fusion scheme, while Liu et al. \cite{liu2018efficient} extend it to the generic multi-view case by low-rank decomposition. 
Unlike multi-view fusion, multi-view alignment intends to align the learned representations from each view, so as to exploit the common information among different views. 
The most popular and representative multi-view alignment method is canonical correlation analysis (CCA) \cite{hotelling1992relations} and its deep variant deep CCA (DCCA) \cite{andrew2013deep}, which seeks to maximize the correlation of two views. 
Wang et al. \cite{wang2015deep} later develop a variant named deep canonically correlated auto-encoders (DCCAE), which is regularized by the reconstruction objective, and Benton et al. \cite{benton2019deep} propose deep generalized CCA (DGCCA) to handle with the case of more than two views. 
In addition to correlation, deep multi-view alignment also leverages other metrics. 
For example, Frome et al. \cite{frome2013devise} maximize the dot-product similarity by a hinge rank loss, while Feng et al. \cite{feng2014cross} minimize the $l_2$-norm distance between the learned representations of two views. 
Besides, inspired by GANs, adversarial training is also borrowed to improve multi-view representation learning by learning modality-invariant representations \cite{wang2017adversarial} or cross-view transformation \cite{gu2018look}. 
Consequently, many solutions have been proposed for multi-view deep learning, and they are widely adopted to serve many tasks, such as action recognition, sentiment analysis and image captioning. 
However, none of those works has considered the marriage of OCC and multi-view deep learning, which is exactly the motivation of this paper.

\begin{figure*}[ht]
\centering
\includegraphics[width=\linewidth]{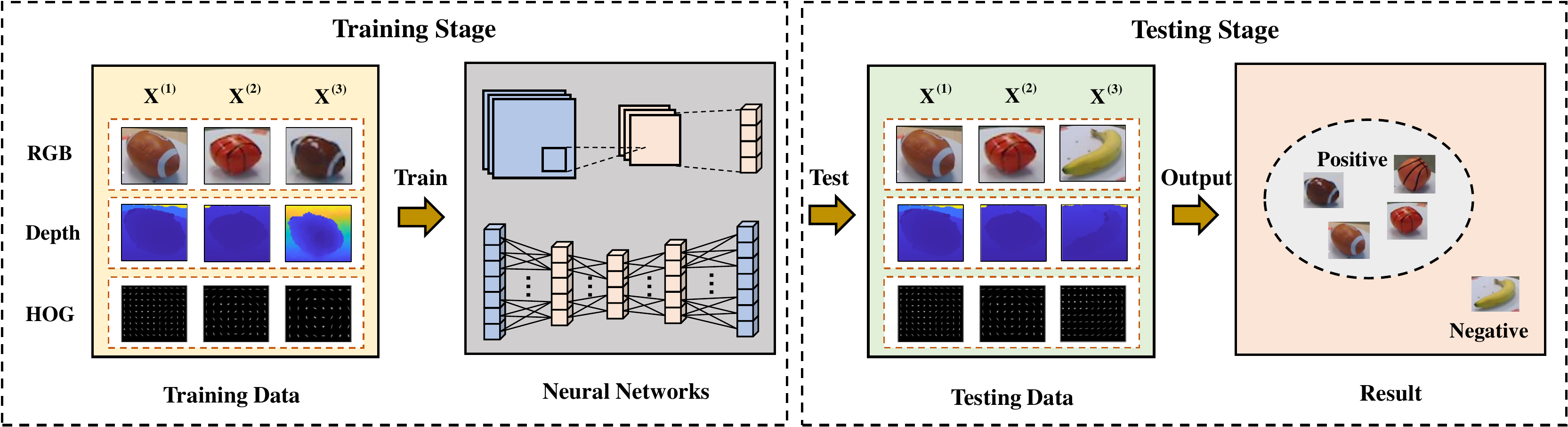}
\caption{An overview of multi-view deep OCC. The example shows soccer balls described by data from three views (RGB, depth and HOG), and the goal is to determine whether incoming multi-view data are from the soccer ball class  or not.}
\label{fig:mvdooc_framework}
\end{figure*}

\section{Problem Formulation}
\label{sec:problem_definition}

To tackle the first obstacle mentioned in Sec. \ref{sec:introduction}, we will provide a formal problem formulation of multi-view deep OCC in the first place. Given the positive class $\mathcal{C}_t$, a multi-view datum $\{\mathbf{x}_{train}^{(v)}\}_{v=1}^V$ is sampled from $\mathcal{C}_t$ for training, where $V\geq 2$ is the number of views and $\mathbf{x}_{train}^{(v)}\in \mathbb{R}^{d^{(v)}_1\times d^{(v)}_2\times\cdots d^{(v)}_{M_v}}$ is a $M_v$-dimensional tensor with the shape $d^{(v)}_1\times d^{(v)}_2\times\cdots d^{(v)}_{M_v}$. To be more specific, $\mathbf{x}_{train}^{(v)}$ denotes the observation from the $v_{th}$ view, while $M_v=1$ and $M_v>1$ corresponds to tabular data and complex data (e.g. images or videos) respectively. Note that the observations from different views can be heterogeneous. With the training datum $\{\mathbf{x}_{train}^{(v)}\}_{v=1}^V$, the goal of multi-view deep OCC is to obtain a deep neural network (DNN) based model: 

\begin{equation}
    \mathcal{M}_{\boldsymbol{\Theta}}: \mathbb{R}^{\prod^{M_1}_{i=1}d^{(1)}_i}\times\mathbb{R}^{\prod^{M_2}_{i=1}d^{(2)}_i}\cdots\times\mathbb{R}^{\prod^{M_V}_{i=1}d^{(V)}_i} \mapsto \{0, 1\}
\end{equation}
where $\boldsymbol{\Theta}$ represents the set of all learnable parameters for the model $\mathcal{M}_{\boldsymbol{\Theta}}$. In the inference phase, $\mathcal{M}_{\boldsymbol{\Theta}}$ aims to classify whether an incoming multi-view testing datum $\{\mathbf{x}_{test}^{(v)}\}_{v=1}^V$ belongs to the target distribution $\mathcal{C}_t$ or not, where $\mathbf{x}_{test}^{(v)}\in \mathbb{R}^{d^{(v)}_1\times d^{(v)}_2\times\cdots d^{(v)}_{M_v}}$ denotes the datum from the $v_{th}$ view, i.e.: 

\begin{equation}
    \mathcal{M}_{\boldsymbol{\Theta}}(\{\mathbf{x}_{test}^{(v)}\}_{v=1}^V) =  \begin{cases}
                                                                     \;1, & \text{if}\quad \{\mathbf{x}_{test}^{(v)}\}_{v=1}^V \in \mathcal{D}_p\\
                                                                     \;-1, & \text{if}\quad \{\mathbf{x}_{test}^{(v)}\}_{v=1}^V \notin \mathcal{D}_p.
                                                                    \end{cases}
 \end{equation}
In practice, $\mathcal{M}_{\boldsymbol{\Theta}}$ is usually supposed to obtain a score $\mathcal{S}(\{\mathbf{x}_{test}^{(v)}\}_{v=1}^V)$, which indicates the likelihood that $\{\mathbf{x}_{test}^{(v)}\}_{v=1}^V$ belongs to the positive class. A threshold can be then chosen to binarize the score into the final decision of $\mathcal{M}_{\boldsymbol{\Theta}}$. It should be noted that the DNN based model $\mathcal{M}_{\boldsymbol{\Theta}}$ can be constructed by either pure DNNs or a mixture of DNNs and classic OCC models. Since a mixture of DNNs and classic OCC often suffers from some issues, such as the decoupling of representation learning and classification, we will focus on discussing the the model that consists of pure DNNs. In other words, we discuss the case where $\mathcal{M}_{\boldsymbol{\Theta}}$ is able to perform end-to-end one-class classification. An overview of multi-view deep OCC is presented in Fig. \ref{fig:mvdooc_framework}.

\section{The Proposed Baselines}
\label{sec:the_proposed_baselines}

Having provided a formal problem formulation, we will address the second issue in Sec. \ref{sec:introduction} by designing baseline solutions to multi-view deep OCC, so as to provide the first sense to approach this topic. In this section, we systematically design four types of baseline solutions: Fusion based solutions, alignment based solutions, tailored deep one-class OCC and self-supervision based solutions.

\subsection{Fusion based Solutions}


A core issue for multi-view learning is how to maximally exploit the information embedded in different views to perform downstream tasks. To this end, the most straightforward idea is to fuse data from multiple views into a joint embedding. Therefore, it is natural for us to propose fusion based multi-view deep OCC solutions, which aim to fuse the data embeddings learned from different views into a joint embedding for OCC. We will discuss its framework and specific implementations of each component below.

\subsubsection{Framework}
\label{sec:fusion_framework}
Given a multi-view datum $\{\mathbf{x}_{train}^{(v)}\}_{v=1}^V$ with $V$ views for training, fusion based solutions first introduce a set of DNN based encoders to encode the input observation of each view into their latent embeddings. For the $v_{th}$ view, an encoder $Enc^{(v)}: \mathbb{R}^{d^{(v)}_1\times d^{(v)}_2\times\cdots d^{(v)}_{M_v}}\mapsto \mathbb{R}^{d^{(v)}_{l}}$ encodes $\mathbf{x}_{train}^{(v)}$ into a latent embedding $\mathbf{h}^{(v)}$ with a dimension $d^{(v)}_l$:

\begin{equation}\label{eq:encoder_in_fusion}
    \mathbf{h}^{(v)} = Enc^{(v)}(\mathbf{x}_{train}^{(v)}),\quad v=1, 2\cdots, V
\end{equation}
where $\mathbf{h}^{(v)}$ is a $d^{(v)}_l$-dimensional column vector. In this way, embeddings from different views can be collected as a set $\{\mathbf{h}^{(v)}\}^V_{v=1}$. Subsequently, fusion based methods select a fusion function $F_f: \mathbb{R}^{d^{(1)}_l\times d^{(2)}_l\times\cdots d^{(V)}_{l}}\mapsto \mathbb{R}^{D}$ to fuse the embeddings of different views into a $D$-dimensional vector ${\mathbf{h}}$ as the joint embedding of the multi-view datum:

\begin{equation}\label{eq:fusion_module_in_fusion}
    {\mathbf{h}} = F_f(\{\mathbf{h}^{(v)}\}_{v=1}^V)
\end{equation}
Since only very weak supervision is available (i.e. all training data share a common positive label), discriminative information is unavailable for guiding the representation learning of encoders in multi-view deep OCC. Therefore, as a baseline, we propose to leverage the frequently-used reconstruction paradigm to guide the model training. To this end, a set of DNN based decoders are introduced to decode the input data of each view from the joint embedding $\mathbf{h}$: For the $v_{th}$ view, an decoder $Dec^{(v)}: \mathbb{R}^{D} \mapsto \mathbb{R}^{d^{(v)}_1\times d^{(v)}_2\times\cdots d^{(v)}_{M_v}}$ intends to map $\mathbf{h}$ back to $v_{th}$ view's original input ${\mathbf{x}}_{train}^{(v)}$:

\begin{equation}\label{eq:decoder_in_fusion}
    \hat{\mathbf{x}}_{train}^{(v)} = Dec^{(v)}({\mathbf{h}}).
\end{equation}
where $\hat{\mathbf{x}}_{train}^{(v)}$ is the reconstructed input of $v_{th}$ view. To train the DNN based model, one can simply minimize the differences between original inputs and reconstructed inputs:

\begin{equation}\label{eq:recon_loss}
    \mathcal{L}_r = \sum_{v=1}^V \ell(\mathbf{x}_{train}^{(v)},\hat{\mathbf{x}}_{train}^{(v)})= \sum_{v=1}^V \vert\vert\mathbf{x}_{train}^{(v)}-\hat{\mathbf{x}}_{train}^{(v)}\vert\vert^2_2
\end{equation}

\begin{figure}[ht]
\centering
\includegraphics[width=0.95\linewidth]{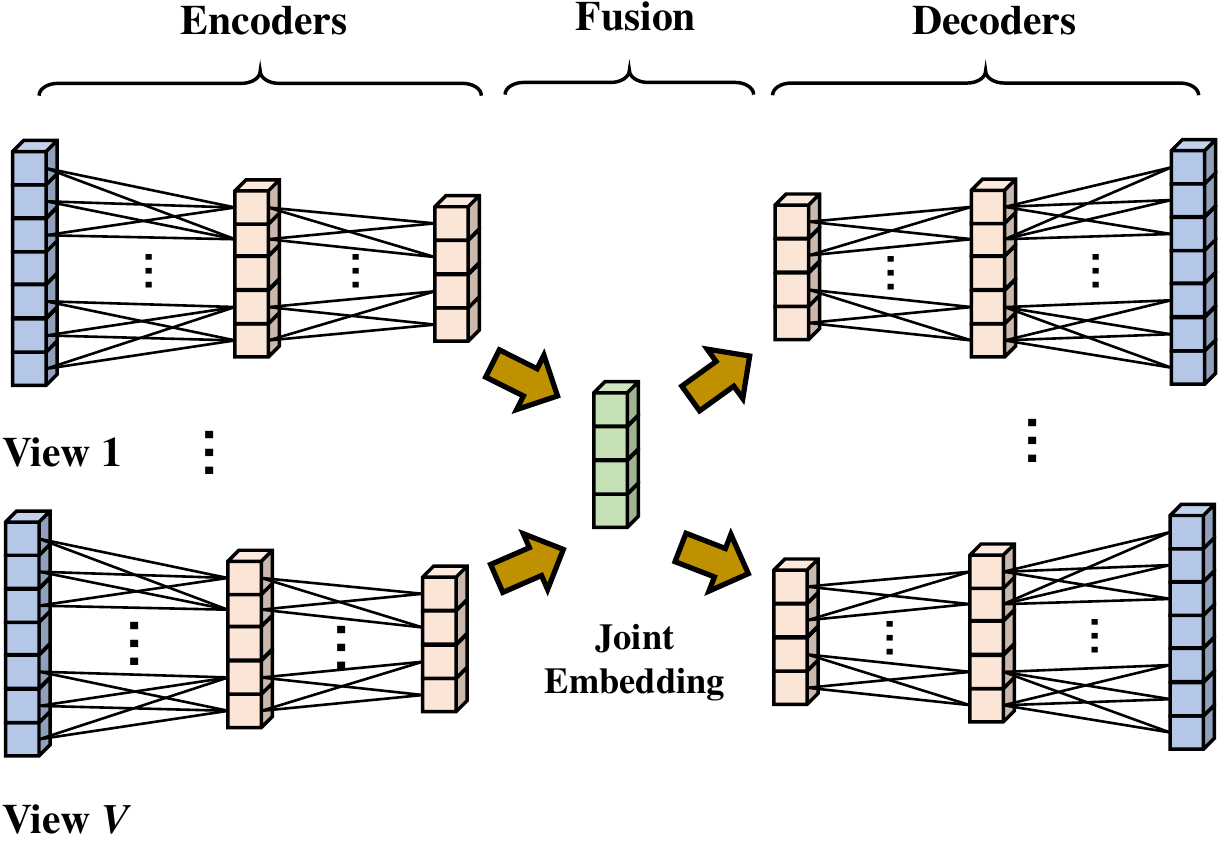}
\caption{Fusion based solutions for multi-view deep OCC.}
\label{fig:mvdooc_fusion}
\end{figure}

In addition to the commonly-used mean square errors (MSE) above, other types of reconstruction loss $\ell(\cdot)$ are also applicable, such as $L_1$-norm reconstruction loss. During testing, an incoming multi-view datum $\{\mathbf{x}_{test}^{(v)}\}_{v=1}^V$ is fed into the network to obtain the reconstructed datum $\{\hat{\mathbf{x}}_{test}^{(v)}\}_{v=1}^V$ by Eq. \ref{eq:encoder_in_fusion} - \ref{eq:decoder_in_fusion}. As the DNN based model is trained with only data from the positive class $\mathcal{C}_t$, one can follow the standard practice in OCC to assume that lower reconstruction errors indicate a higher likelihood that the testing datum belongs to $\mathcal{C}_t$. In other words, a baseline score for the $v_{th}$ view can be directly obtained by $\mathcal{S}^{(v)}(\mathbf{x}^{(v)}_{test})=-\ell(\mathbf{x}_{train}^{(v)},\hat{\mathbf{x}}_{train}^{(v)})$. Finally, we can obtain a score function by the reconstruction errors of all views:
\begin{equation}\label{eq:score}
    \mathcal{S}(\{\mathbf{x}_{test}^{(v)}\}_{v=1}^V) = F_l(\mathcal{S}^{(1)}(\mathbf{x}^{(1)}_{test}),\cdots,\mathcal{S}^{(V)}(\mathbf{x}^{(V)}_{test}))
\end{equation}
where $F_l(\cdot)$ is a late fusion function that combines scores of different views into a final score, which is discussed later. An intuitive illustration of the framework is given in Fig. \ref{fig:mvdooc_fusion}.

\subsubsection{Implementations}\label{sec:fusion_func}

The key to realize a fusion based multi-view deep OCC method is the implementation of fusion function $F_f(\cdot)$. Thus, we design five specific ways to realize $F_f(\cdot)$:


1) \textit{Summation based fusion} (abbreviated as SUM, and the abbreviation of other methods are similarly given). Summation based fusion combines latent embeddings from different views by summing them up. Specifically, it assumes that embeddings of all views share the same dimension $d_l^{(v)}=D$, and a joint embedding $\mathbf{h}$ can be yielded by:
\begin{equation}
     F_f(\{\mathbf{h}^{(v)}\}_{v=1}^V)=\frac{1}{V}\sum_{v=1}^V\mathbf{h}^{(v)}
\end{equation}
When the dimensions of embeddings are different, we can introduce a linear mapping parameterized by a learnable matrix $\mathbf{P}^{(v)}$ to map the $v_{th}$ embedding to the shared dimension $D$: $\textstyle\hat{\mathbf{h}}^{(v)} = \mathbf{P}^{(v)}\cdot\mathbf{h}^{(v)}$. 
Since neural networks can flexibly map a datum into an embedding with any dimension with a linear mapping layer, we simply assume that all embeddings $\mathbf{h}^{(v)}$ share the same dimension $D$ here to facilitate analysis in the rest parts of this paper.

2) \textit{Max based fusion} (MAX). 
Similar to summation based fusion, max based fusion also assumes a shared dimension $d_l^{(v)}=D$ and use the maximum of embeddings of different views as the joint embedding ${\mathbf{h}}$:

\begin{equation}
    F_f(\{\mathbf{h}^{(v)}\}_{v=1}^V)=\max(\{\mathbf{h}^{(v)}\}_{v=1}^V)
\end{equation}


3) \textit{Network based fusion} (NN). It is easy to notice that both summation based fusion and max based fusion assume a shared embedding dimension across different views. To make the fusion more flexible, it is also natural to map all latent embeddings into the joint embedding ${\mathbf{h}}$ by a fully-connected neural network with learnable parameters: 
\begin{equation}
    F_f(\{\mathbf{h}^{(v)}\}_{v=1}^V) = \sigma(\mathbf{W}\cdot Cat(\{\mathbf{h}^{(v)}\}_{v=1}^V) + \mathbf{b})
\end{equation}
where $\mathbf{W}$ and $\mathbf{b}$ are learnable weights and biases of corresponding neurons, and $Cat(\cdot)$ and $\sigma(\cdot)$ denote the concatenation operation and the activation function respectively. Note that we can also leverage a multi-layer fully-connected network to perform DNN based fusion.

4) \textit{Tensor based fusion} (TF). Tensor based fusion \cite{zadeh2017tensor} is an emerging method in multi-view deep learning. The core idea of tensor based fusion is to combine the embeddings of different views by the tensor outer product $\textstyle\mathcal{Z} = \bigotimes_{v=1}^V \mathbf{h}^{(v)}$, where $\mathcal{Z}$ is a $d^{(1)}_l\times d^{(2)}_l\times\cdots d^{(V)}_{l}$ tensor. Afterwards, $\mathcal{Z}$ is fed into a linear layer with weight tensor $\mathcal{W}\in \mathbb{R}^{D\times d^{(1)}_l\times d^{(2)}_l\times\cdots d^{(V)}_{l}}$ and bias vector $b\in \mathbb{R}^D$ to obtain the unified representation $\mathbf{h}$:
\begin{equation}
    F_f(\{\mathbf{h}^{(v)}\}_{v=1}^V) = \mathcal{W}\cdot \mathcal{Z}+b
\end{equation}
Note here we slightly abuse the notation of matrix-vector multiplication by considering $\mathcal{W}$ and $\mathcal{Z}$ as $D\times K$ matrix and $K$-dimensional vector, where $K=\prod^V_{v=1}d^{(v)}_{l}$. However, a severe practical problem is that tensor based fusion requires computing the tensor $\mathcal{Z}$ and recording $\mathcal{W}$, which incurs exponential computational cost. To address this problem, we leverage the low-rank approximation technique in \cite{liu2018efficient} by considering the calculation of the unified representation $\mathbf{h}$'s $k_{th}$ element, $\mathbf{h}(k)$. Suppose that the weight $\mathcal{W}$ is yielded by stacking $D$ tensors $\mathcal{W}=[\mathcal{W}_1;\mathcal{W}_2\cdots;\mathcal{W}_D]$, where $\mathcal{W}_k\in \mathbb{R}^{1\times d^{(1)}_l\times d^{(2)}_l\times\cdots d^{(V)}_{l}}$ and $k=1,\cdots,D$. Thus, we have:

\begin{equation}
    \mathbf{h}(k) = \mathcal{W}_k\cdot \mathcal{Z}+b(k)
\end{equation}
where $b(k)$ is the $k_{th}$ element of $b$. Then, $\mathcal{W}_k$ can be approximated by a set of learnable vectors as follows:
\begin{equation}
 \mathcal{W}_k = \sum_{r=1}^R \bigotimes_{v=1}^V\mathbf{w}_{r,k}^{(v)}
\end{equation}
where $\mathbf{w}_{r,k}^{(v)}\in \mathbb{R}^{d^{(v)}_l}$ and $R$ is the rank of low-rank approximation. Since $\mathcal{Z}=\bigotimes_{v=1}^V \mathbf{h}^{(v)}$, tensor based fusion can be computed in a highly efficient manner by rearranging the order of inner product and outer product \cite{liu2018efficient}, which enables tensor based fusion to be computationally tractable.




\subsection{Alignment based Solutions}

Compared with multi-view fusion, multi-view alignment is another popular category of methods in multi-view deep learning. It does not require to obtain a joint embedding. Instead, they attempt to align the representations learned by different views, so as to make those representations share some common characteristics. Likewise, we also present the overall framework and specific implementations of alignment based multi-view deep OCC solutions below.


\subsubsection{Framework}
In a training batch with $N$ multi-view data, we denote the embeddings of the $n_{th}$ multi-view datum $\{\mathbf{x}^{(v)}_{n}\}^{V}_{v=1}$ by $\{\mathbf{h}_n^{(v)}\}^V_{v=1}$, which are learned by a set of encoder networks $\{Enc^{(v)}\}^V_{v=1}$. Then, an alignment function $F_a$ is defined to compute a quantitative measure of alignment across learned embeddings of different views:

\begin{equation}\label{eq:align}
    \mathcal{A} = F_a(\{\{\mathbf{h}_n^{(v)}\}^V_{v=1}\}_{n=1}^N),\quad F_a\in \mathcal{F}_a
\end{equation}
where $\mathcal{F}_a$ is the set of available alignment functions. As shown in Eq. \ref{eq:align},  a key difference between alignment based solutions and fusion based solutions is that fusion usually occurs within one multi-view datum, while the alignment of two views can involve multiple multi-view data. To maximize the alignment across different views, we can equivalently minimize the alignment loss $\mathcal{L}_a=-\mathcal{A}$. Similar to fusion based solutions, we also resort to the reconstruction paradigm and a set of decoder networks $\{Dec^{(v)}\}^V_{v=1}$ to guide the training of DNNs. As a result, alignment based solutions minimize the following loss function:

\begin{equation}
    \mathcal{L} = \mathcal{L}_r + \alpha\; \mathcal{L}_a,
\end{equation}
where $\mathcal{L}_r$ is the reconstruction loss defined in Eq. \ref{eq:recon_loss} and $\alpha$ is the weight of alignment loss. Given a testing datum $\{\mathbf{x}_{test}^{(v)}\}_{v=1}^V$, we also leverage the reconstruction errors as baseline scores, which is the same as Eq. \ref{eq:score}.


\begin{figure}[t]
\centering
\includegraphics[width=0.95\linewidth]{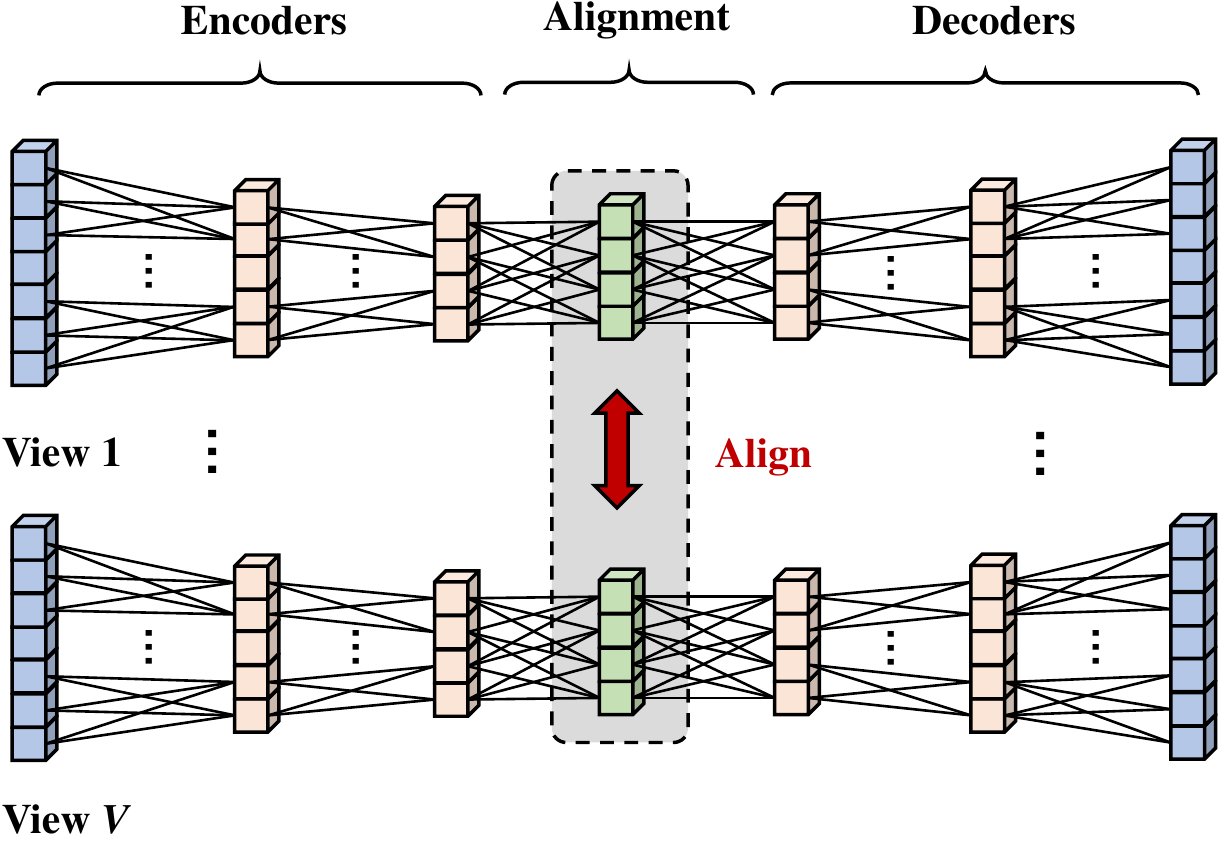}
\caption{Alignment based solutions for multi-view deep OCC.}
\label{fig:mvdooc_alignment}
\end{figure}

\subsubsection{Implementations}
The core issue for alignment based solutions is the design of alignment function $F_a$. Inspired by the literature of multi-view deep learning, we propose to implement the alignment function by the following ways:

1) \textit{Distance based alignment} (DIS). 
A commonly-seen technique to align two embeddings is to minimize their distance, as a smaller distance usually indicates better alignment. Therefore, we propose to adopt the widely-used pair-wise $L_p$-norm distance of all embeddings to measure alignment:

\begin{equation}\label{eq:dis_align}
    F_a(\{\{\mathbf{h}_n^{(v)}\}^V_{v=1}\}_{n=1}^N) = -\sum^N_{n=1}\sum_{i=1}^{V-1}\sum_{j=i+1}^{V} \vert\vert\mathbf{h}_n^{(i)}-\mathbf{h}_n^{(j)}\vert\vert^p_p
\end{equation}
where $\vert\vert\cdot\vert\vert$ denotes the $L_p$-norm and $p$ is a non-negative integer. As can be seen from Eq. \ref{eq:dis_align}, it requires the embeddings of different views to share a common dimension. Note that the alignment function in Eq. \ref{eq:dis_align} is equivalent to the correspondent autoencoder proposed in \cite{feng2014cross}, which leverages multi-view alignment for cross-modal retrieval. A drawback of such an alignment function is that it only performs the view alignment within one multi-view datum.

2) \textit{Similarity based alignment} (SIM). 
In addition to distance, similarity is another intuitive way to measure the degree of alignment. Given a similarity function $s(\cdot)$, we can similarly define an alignment function like Eq. \ref{eq:dis_align}. Nevertheless, such an alignment function only considers the view similarity within one multi-view datum. To consider the view similarity across different datum, we are inspired by \cite{frome2013devise} and propose to adopt a more sophisticated similarity measure between different views: For $i_{th}$ and $j_{th}$ view, a similarity loss $Sim(i,j)$ is computed based on $s(\cdot)$ and a hinge loss:

\begin{equation}
    Sim(i,j)=\sum_{a\neq b}\max\{0, m-s(\mathbf{h}_a^{(i)}, \mathbf{h}_a^{(j)})+s(\mathbf{h}_a^{(i)}, \mathbf{h}_b^{(j)})\},
\end{equation}
where $m$ is a margin. The above similarity loss encourages the embeddings from the same multi-view datum to be similar, while embeddings from two different different multi-view data to be dissimilar. The similarity function $s(\cdot)$ can be realized by multiple forms, such as inner product and cosine similarity. Then, the final alignment function can be calculated by:

\begin{equation}
    F_a(\{\{\mathbf{h}_n^{(v)}\}^V_{v=1}\}_{n=1}^N) = \sum_{i=1}^{V-1}\sum_{j=i+1}^{V}Sim(i,j)
\end{equation}

3) \textit{Correlation based alignment} (DCCA). 
Canonical correlation analysis (CCA) is a classic statistical technique for finding the maximally correlated linear projections of two vectors.
Thus, a natural way for us to align different views in deep learning is the CCA's deep variant, deep CCA \cite{andrew2013deep}.
To conduct correlation based alignment, we intend to maximize the correlation between two views. Specifically, we stack $v_{th}$ view's embeddings of $N$ multi-view data in a training batch into a $d^{(v)}_l\times N$ embedding matrix: $\textstyle\mathbf{H}^{(v)}=[\mathbf{h}^{(v)}_1,\cdots,\mathbf{h}^{(v)}_N]$, while $\mathbf{H}^{(v)}$ can be centered by $\textstyle\bar{\mathbf{H}}^{(v)}=\mathbf{H}^{(v)}-\frac{1}{N}\mathbf{H}^{(v)}\cdot\mathbf{1}$, where $\mathbf{1}$ is a $N\times d^{(v)}_l$ all-1 matrix. With the embedding matrix $\mathbf{H}^{(i)}$ and $\mathbf{H}^{(j)}$ for the $i_{th}$ and $j_{th}$ view, we first estimate the covariance matrices $\textstyle \sum_{ii}=\frac{1}{N-1}\bar{\mathbf{H}}^{(i)}\cdot \bar{\mathbf{H}}^{(i)\top}+r\mathbf{I} $, $\textstyle \sum_{ij}=\frac{1}{N-1}\bar{\mathbf{H}}^{(i)}\cdot \bar{\mathbf{H}}^{(j)\top}$ and $\textstyle \sum_{jj}=\frac{1}{N-1}\bar{\mathbf{H}}^{(j)}\cdot \bar{\mathbf{H}}^{(j)\top}+r\mathbf{I}$, 
where $r$ is the coefficient for regularization and $\mathbf{I}$ is an identity matrix. With estimated covariance matrices, we compute an intermediate matrix $T_{ij}=\textstyle \sum^{-{1}/{2}}_{ii}\cdot \sum_{ij}\cdot \sum^{-{1}/{2}}_{jj}$. It can be proved that the correlation of view $i$ and $j$ is the matrix trace norm of $T_{ij}$ \cite{andrew2013deep}:

\begin{equation}
    Corr(i, j) = \vert\vert T_{ij}\vert\vert=tr(T_{ij}\cdot T^\top_{ij})^{\frac{1}{2}}
\end{equation}

The final alignment function can be calculated by:

\begin{equation}
    F_a(\{\{\mathbf{h}_n^{(v)}\}^V_{v=1}\}_{n=1}^N) = \sum_{i=1}^{V-1}\sum_{j=i+1}^{V}Corr(i,j)
\end{equation}

\subsection{Tailored Deep OCC Solutions}

Apart from baselines based on multi-view deep learning, we design the third type of baseline solutions by tailoring existing deep OCC solutions. The basic idea is to train a deep OCC model for data from each view. During inference, the OCC results of each view are fused to yield the final results. The framework and specific implementations of tailored deep OCC solutions are presented below.


\subsubsection{Framework}

Suppose that the deep OCC model $\mathcal{M}^{(v)}$ is trained with data from the $v_{th}$ view. Given a newly incoming multi-view datum $\{\mathbf{x}^{(v)}_{test}\}^V_{v=1}$, the OCC result for the $v_{th}$ view is given by:

\begin{equation}
    \mathcal{S}^{(v)} = \mathcal{M}^{(v)}(\mathbf{x}^{(v)}_{test}),\quad v=1,\cdots,V
\end{equation}
The final score for the multi-view datum $\{\mathbf{x}^{(v)}_{test}\}^V_{v=1}$ is computed by a late fusion function $F_l(\cdot)$:
\begin{equation}
    \mathcal{S}(\{\mathbf{x}_{test}^{(v)}\}_{v=1}^V) = F_l(\mathcal{S}^{(1)}(\mathbf{x}^{(1)}_{test}), \mathcal{S}^{(2)}(\mathbf{x}^{(2)}_{test})\cdots,\mathcal{S}^{(V)}(\mathbf{x}^{(V)}_{test}))
\end{equation}

\begin{figure}[t]
\centering
\includegraphics[width=0.95\linewidth]{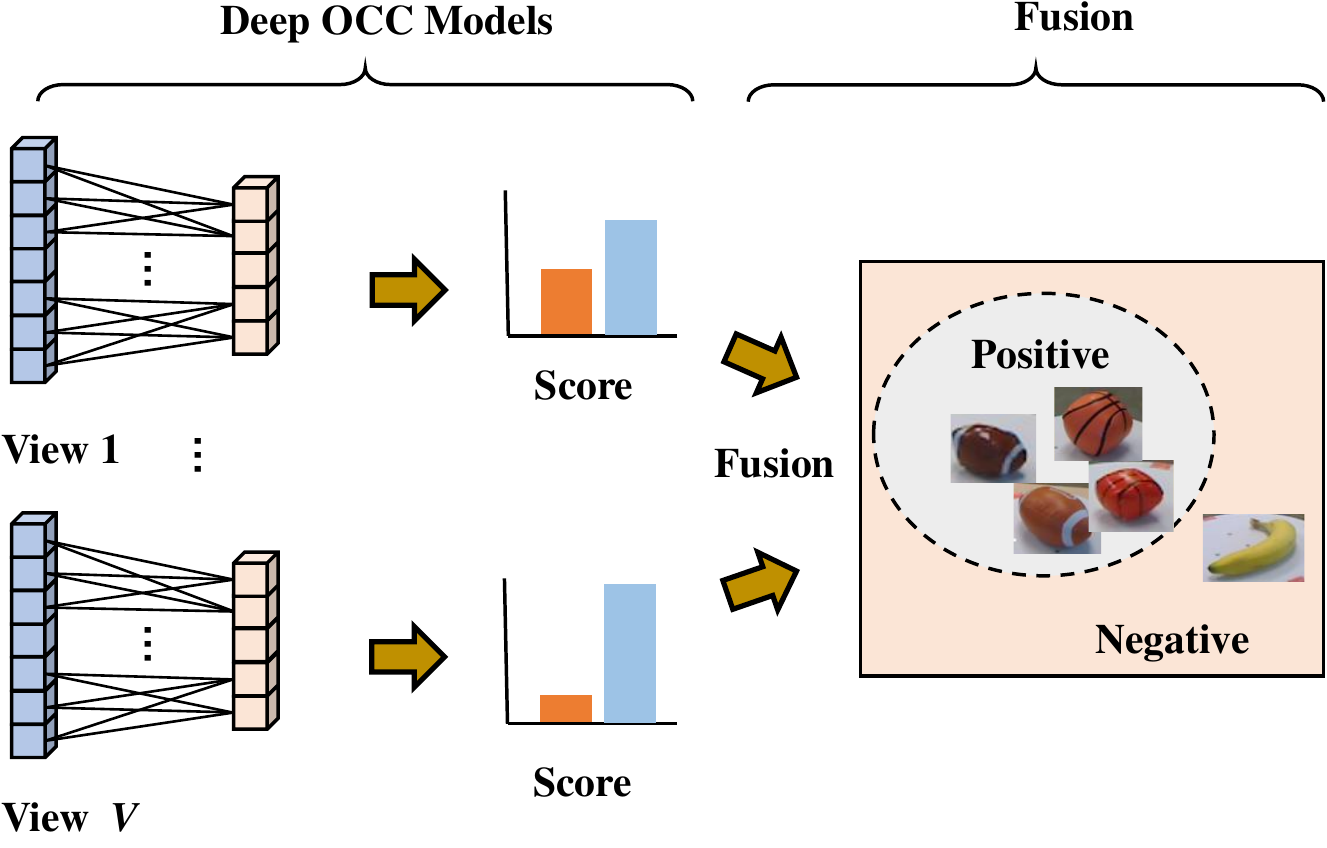}
\caption{Framework of tailored deep OCC solutions.}
\label{fig:mvdooc_docc}
\end{figure}


\subsubsection{Implementations}


The choice of deep OCC model plays a center role in designing tailored deep OCC solutions. This paper introduces two representative deep OCC methods in the literature to construct baseline models for multi-view deep OCC: Standard deep autoencoders (DAE) and the recent deep support vector data description (DSVDD) \cite{ruff2018deep}:

1) \textit{DAE based solution} (DAE). DAE leverages DNNs as the encoder $Enc^{(v)}$ and the decoder $Dec^{(v)}$ to reconstruct input data from a low-dimensional embedding. Formally, given $N$ training data $\{\mathbf{x}^{(v)}_n\}^N_{n=1}$ from the $v_{th}$ view, DAE requires to minimize the reconstruction loss:

\begin{equation}
\begin{split}
    \min_{\boldsymbol{\theta}^{(v)}_E, \boldsymbol{\theta}^{(v)}_D}\quad &\frac{1}{N}\sum^N_{n=1}\vert\vert Dec^{(v)}(Enc^{(v)}(\mathbf{x}^{(v)}_n))-\mathbf{x}^{(v)}_n\vert\vert^2_2 \\
     & +\frac{\lambda}{2}(\vert\vert\boldsymbol{\theta}^{(v)}_E\vert\vert^2_2+\vert\vert\boldsymbol{\theta}^{(v)}_D\vert\vert^2_2) 
\end{split}
\end{equation}
where $\boldsymbol{\theta}^{(v)}_E$ and $\boldsymbol{\theta}^{(v)}_D$ represent the learnable parameters of the encoder network $Enc^{(v)}$ and decoder network $Dec^{(v)}$ respectively, and $\lambda$ is the weight for the $L_2$-norm regularization term. For inference, the reconstruction errors are often directly used as scores:

\begin{equation}\label{eq:simple_dsvdd}
    \mathcal{S}^{(v)} = \vert\vert Dec^{(v)}(Enc^{(v)}(\mathbf{x}^{(v)}_{test}))-\mathbf{x}^{(v)}_{test}\vert\vert_2^2
\end{equation}

DAE based baseline is also viewed as the most fundamental baseline for multi-view deep OCC.

2) \textit{DSVDD based solution} (DSV). In general, DSVDD intends to map embeddings of data from the positive class $\{\mathbf{h}^{(v)}_n\}^N_{n=1}$ to a hyper-sphere with minimal radius. To be more specific, DSVDD can be implemented by a simplified version or a soft-boundary version \cite{ruff2018deep}. Since the simplified version enjoys less hyperparameters and better performance in practice, we choose it to perform multi-view deep OCC. Specifically, {simplified DSVDD} encourages embeddings of all data to be as close to a center $\mathbf{c}^{(v)}$ as possible. Formally, simplified DSVDD requires to solve the following optimization problem: 

\begin{equation}\label{eq:simple_dsvdd}
    \min_{\boldsymbol{\theta}_E^{(v)}}\quad \frac{1}{N}\sum^N_{n=1}\vert\vert Enc^{(v)}(\mathbf{x}^{(v)}_n))-\mathbf{c}^{(v)}\vert\vert^2_2+\frac{\lambda}{2}\vert\vert\boldsymbol{\theta}_E^{(v)}\vert\vert^2_2 
\end{equation}
where $\boldsymbol{\theta}^{(v)}$ are the learnable parameters of DSVDD, and $\lambda$ is the weight for the $L_2$-norm regularization term. The encoder $Enc^{(v)}$ can be pre-trained in a DAE fashion. The non-zero center $\mathbf{c}^{(v)}$ is initialized before training and can be adjusted during training. The above optimization problem can be efficiently solved by gradient descent.  During inference, one can score a test datum $\mathbf{x}^{(v)}_{test}$ by calculating the distance between its embedding $\mathbf{h}_{test}^{(v)}$ and the center:

\begin{equation}\label{eq:simple_dsvdd}
    \mathcal{S}^{(v)} = \vert\vert Enc^{(v)}(\mathbf{x}^{(v)}_{test}))-\mathbf{c}^{(v)}\vert\vert_2^2
\end{equation}

\subsection{Self-supervision based Solutions}

Self-supervised learning is a hot topic in recent research, and it has been demonstrated as a highly effective way to conduct unsupervised representation learning \cite{jing2020self}. Specifically, self-supervised learning introduces a certain pretext task to provide additional supervision signal and enable better representation learning. Due to the lack of supervision signal in multi-view deep OCC, creating self-supervision can be an appealing solution. Multi-view data intrinsically contains richer information than single-view data, which makes it possible to design pretext tasks in a more flexible way. In this section, we mainly focus on designing generative pretext tasks to realize self-supervised multi-view deep OCC. We also explore discriminative pretext tasks in the supplementary material.

\subsubsection{Framework}

The basic intuition for generative pretext tasks is to generate data from some views based on data from other views. Formally, given a multi-view datum $\{\mathbf{x}^{(v)}\}_{v=1}^V$, we partition the view indices into two subsets $\mathcal{P}$ and $\mathcal{Q}$, which satisfy:

\begin{equation}
    \mathcal{P}\neq\mathcal{Q},\quad \mathcal{P}\cup\mathcal{Q}=\{1,2,\cdots,V\} 
\end{equation}
Note that the intersection of $\mathcal{P}$ and $\mathcal{Q}$ may not be empty. By $\mathcal{P}$ and $\mathcal{Q}$, we can partition the multi-view data into two sets of data, $\{\mathbf{x}^{(i)}\}_{i\in \mathcal{P}}$ and $\{\mathbf{x}^{(j)}\}_{j\in \mathcal{Q}}$. The goal of generative pretext tasks is to generate $\{\mathbf{x}^{(j)}\}_{j\in \mathcal{Q}}$ by taking $\{\mathbf{x}^{(i)}\}_{i\in \mathcal{P}}$ as inputs. To fulfill this task, we propose to introduce $\vert\mathcal{P}\vert$ encoder networks and $\vert\mathcal{Q}\vert$ decoder networks, where $\vert\cdot\vert$ denotes the number of elements in the set. $\{\mathbf{x}^{(i)}\}_{i\in \mathcal{P}}$ is first mapped to the embedding set $\{\mathbf{h}^{(i)}\}_{i\in \mathcal{P}}$ by encoders, and a joint embedding is then obtained by:

\begin{equation}
    \mathbf{h}^{\mathcal{P}}=F_f(\{\mathbf{h}^{(i)}\}_{i\in \mathcal{P}})
\end{equation}
where $F_f(\cdot)$ can be any fusion function defined in Sec. \ref{sec:fusion_func}. The decoder networks use $\mathbf{h}^{\mathcal{P}}$ as the input to infer the data $\{\mathbf{x}^{(j)}\}_{j\in \mathcal{Q}}$, which aims to learn:

\begin{equation}\label{eq:simple_dsvdd}
    \min_{\boldsymbol{\theta}_D^{(j)}}\quad \sum_{j\in \mathcal{Q}}\vert\vert Dec^{(j)}(\mathbf{h}^{\mathcal{P}})-\mathbf{x}^{(j)}\vert\vert^2_2
\end{equation}
where $\boldsymbol{\theta}_D^{(j)}$ is the set of learnable weights for decoder $Dec^{(j)}$. In this way, $\{\mathbf{x}^{(j)}\}_{j\in \mathcal{Q}}$ is used as supervision signal to guide the training of encoders and decoders. Similarly, the generation errors $\ell(Dec^{(j)}(\mathbf{h}^{\mathcal{P}}), \mathbf{x}^{(j)})$ can be used for scoring during inference. 

\subsubsection{Implementations}
There are many ways to divide $\mathcal{P}$ and $\mathcal{Q}$, we select two of them to build our baseline solutions here: 

1) \textit{Plain prediction} (PPRD), where $\mathcal{Q}=\{v\}$ and  $\mathcal{P}=\{1,2,\cdots,V\}-\mathcal{Q}$. It means that we predict data from the $v_{th}$ view by data from the rest of views. Due to the lack of standard to select a specific $v$, we vary $v$ from $1$ to $V$, and alternatively use each view as learning target, which results in multiple rounds of prediction. To avoid excessive computational cost, we introduce $V$ encoders and $V$ decoders in total, and the $v_{th}$ encoder and decoder are specifically responsible for data of the $v_{th}$ view in each round of prediction. The final score is yielded by averaging the results of all rounds of prediction. 

2) \textit{Split prediction} (SPRD), where $\mathcal{P}=\{v\}$ and $\mathcal{Q}=\{1,2,\cdots,V\}$. It means that we predict data of all views by a data from the $v_{th}$ view. Likewise, we also alternatively use data of each view to predict data of all views, and introduce $V$ encoders and $V$ decoders that are shared in different rounds of prediction. As shown above, generative pretext tasks aim to maximally capture the inter-view correspondence during representation learning, which cannot be realized by previous baseline solutions.

\subsection{Additional Remarks}
1) \textit{Late Fusion}. 
Except for the self-supervision based solution that uses discriminative pretext tasks, all other baselines require to fuse the results yielded by different views via a late fusion function $F_l$. For traditional tasks like classification and clustering \cite{liu2013sample,liu2018late}, numerous strategies have been proposed to carry out late fusion. However, since OCC lacks discriminative supervision information and trains the model with only data from the positive class, it is not straightforward to exploit prior knowledge or propose an assumption on different views to perform late fusion. Thus, considering that the average strategy is usually viewed as a non-trivial baseline in traditional multi-view learning \cite{liu2021multiview,DBLP:conf/aaai/JiyuanAAAI21Hierarchical}, we also adopt the simple averaging strategy for late fusion in our baseline solutions above, namely:

\begin{equation}
     F_l(\mathcal{S}^{(1)}, \mathcal{S}^{(2)}\cdots,\mathcal{S}^{(V)})=\frac{1}{V}\sum^V_{v=1}\mathcal{S}^{(v)}
\end{equation}
Apart from the simple averaging, one can certainly adopt more sophisticated late fusion strategy, such as the covariance based late fusion strategy proposed in \cite{xu2017detecting}. However, our later empirical evaluations show that late fusion based averaging is a fairly strong baseline, which often prevails in both effectiveness and efficiency. 2) \textit{Other Potential Baselines}. Actually, we have explored more ways to design baseline solutions for multi-view deep OCC. Twelve baseline solutions presented above are the most representative ones that enjoy easier implementation, satisfactory performance and sound extendibility. Due to the limit of pages, we introduce other potential baseline solutions in the supplementary material.

\section{Benchmark Datasets}\label{sec:dataset}

\subsection{Limitations of Existing Datasets}

Public benchmark datasets play an pivotal role in prompting the development of machine learning algorithms, and multi-view deep OCC is no exception. However, as we briefed in Sec. \ref{sec:introduction} (the third issue), existing datasets basically suffer from some important limitations when they are used for evaluating multi-view deep OCC algorithm: 1) \textit{First, existing multi-view datasets are not adequate for multi-view deep OCC.} To be more specific, frequently-used multi-view benchmark datasets (e.g. \textit{Flower17/Flower102}\footnote{\url{https://www.robots.ox.ac.uk/~vgg/data/flowers/17/}}, \textit{BBC}\footnote{\url{http://mlg.ucd.ie/datasets/bbc.html}}) are originally designed to evaluate traditional multi-view learning algorithms, and the data number of them are too small to train DNNs (the average data number of a class is often less than 100). As a consequence, very few existing multi-view datasets can be directly adopted for multi-view deep OCC. 2) \textit{Second, popular benchmark datasets for deep learning are typically single-view}. Recent years have witnessed a surging interest in deep learning, which gives rise to a rapid growth of available benchmark datasets. By contrast, multi-view deep learning is still relatively new area with much less applicable benchmark datasets. 3) \textit{Most importantly, almost no benchmark dataset is speficially designed for the background of multi-view deep OCC}. As we introduced in Sec. \ref{sec:introduction}, multi-view deep OCC actually enjoys broad applications in many realms, such as vision based anomaly detection and fault detection, but the establishment of benchmark datasets in such background is still a blank. In the literature, many works adopt the ``one \textit{v.s.} all'' protocol to convert a binary/multi-class dataset into an OCC dataset, which is non-comprehensive for evaluation of multi-view deep OCC. 

To address those issues above, we need to build new multi-view benchmark datasets for multi-view deep OCC. However, collecting multi-view data from scratch can be expensive and time-consuming, and it takes a long time to obtain sufficient and diverse benchmark datasets in this way. Therefore, our strategy is to extensively collect existing data that come from mature public benchmark datasets, and process them via various means into proper multi-view datasets, so as to construct sufficient benchmark datasets in a highly efficient manner. Collected data and processing techniques will be elaborated below.

\subsection{Building New Multi-view Benchmark Datasets}

We intend to build our new multi-view benchmark datasets based on vision data, which is due to the fact that computer vision is the earliest realm where deep learning is thoroughly studied and successfully applied. Hence, abundant accessible public vision data can be exploited in the realm. Specifically, we process existing data into \textit{image based multi-view datasets} and \textit{video based multi-view datasets}.

\subsubsection{Image based Multi-view Datasets}

Image data are the most fundamental data type in deep learning. We process image data into multi-view data by the following means:

1) \textit{Multiple image descriptors}. Many image descriptors have been proposed to depict different attributes of images, such as texture, color and gradient. Therefore, it is natural to convert a single image into a multi-view data by describing it with different image descriptors. In this paper, we choose several popular image benchmark datasets with comparatively small images (e.g. $32\times 32$ images), which are less complex for image descriptors to depict: \textit{MNIST}\footnote{\url{http://yann.lecun.com/exdb/mnist/}}, 
\textit{FashionMNIST}\footnote{\url{https://github.com/zalandoresearch/fashion-mnist/}}, \textit{CIFAR10}\footnote{\url{https://www.cs.toronto.edu/~kriz/cifar.html}\label{fn:cifar}}, \textit{SVHN}\footnote{\url{http://ufldl.stanford.edu/housenumbers/}},
\textit{CIFAR100}\textsuperscript{\ref{fn:cifar}} and nine image datasets from the \textit{MedMNIST} dataset collection\footnote{\url{https://medmnist.github.io/}}.  To obtain multi-view data, we extract six types of features, i.e. color histogram, GIST, HOG2x2, HOG3x3, LBP and SIFT, which are implemented by a feature extraction Toolbox\footnote{\url{https://github.com/adikhosla/feature-extraction}}.


2) \textit{Multiple pre-trained DNN models}. Classic image descriptors often find it hard to describe high-resolution images effectively. Therefore, to convert high-resolution images to multi-view data, we propose to describe them by multiple pre-trained DNN models with different network architectures. Those DNN models are usually pretrained on a large-scale generic image benchmark dataset like ImageNet \cite{deng2009imagenet}, while different network architectures enable them to acquire image knowledge from different views. We extract the outputs from the penultimate layer of each pretrained DNN model as the representations of the image. For high-resolution image data, we collect image data from the \textit{Cat\_vs\_Dog}\footnote{\url{www.diffen.com/difference/Cat_vs_Dog/}} dataset and \textit{MvTecAD}\footnote{\url{https://www.mvtec.com/company/research/datasets/mvtec-ad/}} dataset collection that contains fifteen datasets. As to DNN architectures, we select VGGNet \cite{Simonyan15}, Inceptionv3 \cite{2016Rethinking}, ResNet34 \cite{he2016deep} and DenseNet121 \cite{huang2017densely} pretrained on ImageNet. In addition, it is worth mention that the \textit{MvTecAD} dataset collection is specifically designed for evaluating OCC models, which makes it even more favorable for multi-view Deep OCC. 


\subsubsection{Video based Multi-view Datasets}

Compared with image data, video data contain both spatial and temporal information, so it is even more natural to transform them into multi-view representation. Since anomaly detection is a representative application of OCC, we simply collect video data from benchmark datasets that are designed for the video anomaly detection (VAD) task \cite{ramachandra2020survey}. To yield video data from a different view, we calculate the optical flow map of each video frame by a pretrained FlowNetv2 model \cite{ilg2017flownet}. In this way, each video frame is represented from the view of both RGB and optical flow, which depict videos by both appearance and motion. Afterwards, we leverage the joint foreground localization strategy from \cite{yu2020cloze}, so as to localize both daily and novel video foreground objects by bounding boxes. Based on those bounding boxes, we can extract both corresponding RGB and optical flow patches from the original video frame and optical flow map respectively, which serve as a two-view representation of each foreground object in videos. Extracted patches are then normalized into the same size ($32\times 32$ patches). As to VAD datasets, we select \textit{UCSDped1/UCSDped2}\footnote{\url{http://www.svcl.ucsd.
edu/projects/anomaly/dataset.htm}}, \textit{Avenue}\footnote{\url{http://www.cse.cuhk.edu.hk/
leojia/projects/detectabnormal/dataset.html}}, \textit{UMN}\footnote{\url{http://
mha.cs.umn.edu/proj_events.shtml\#crowd}} and \textit{ShanghaiTech}\footnote{\url{https://svip-lab.github.io/dataset/campus dataset.html}}. For VAD datasets that provide pixel-level ground-truth mask for abnormal video foreground (\textit{UCSD ped1/ped2}, \textit{Avenue} and \textit{ShanghaiTech}), those patches that are overlapped with any anomaly mask are labelled as $-1$, other patches are labelled as $1$. Although \textit{UMN} dataset does not provide pixel-level mask, its anomalies happen at a certain stage, and all foreground objects exhibit abnormal behavior at that stage. Therefore, we simply label each foreground patch in that stage as $-1$, otherwise labelled as $1$. In this way, we can yield video based multi-view datasets, which are readily applicable to evaluate multi-view deep OCC with real-world application background.  

\subsection{Other Multi-view Benchmark Datasets}
\label{sec:other_datasets}
As a supplement to above multi-view datasets, we also review the literature extensively and select some existing multi-view datasets that are applicable to the evaluation of multi-view deep OCC. Since none of existing multi-view datasets is specifically designed for OCC, we adopt the ``one \textit{v.s.} all'' protocol to evaulate multi-view deep OCC methods on them:  At each round, a certain class of the dataset is viewed as the positive class, while all of other classes are viewed as the negative class. The final OCC performance can be obtained by averaging the performance of all rounds. The selection criterion is that at least one class in the multi-view dataset can provide more than $300$ data for training. With this criterion, we select eleven multi-view datasets: \textit{Citeseer}\footnote{\url{http://lig-membres.imag.fr/grimal/data.html}\label{fn:grimal}}, 
\textit{Cora}\textsuperscript{\ref{fn:grimal}}, 
\textit{Reuters}\textsuperscript{\ref{fn:grimal}}, 
\textit{BBC}\footnote{\url{http://mlg.ucd.ie/datasets/segment.html}}, 
\textit{Wiki}\footnote{\url{http://www.svcl.ucsd.edu/projects/crossmodal/}}, 
\textit{BDGP}\footnote{\url{http://ranger.uta.edu/~heng/Drosophila/}}, 
\textit{Caltech20}\footnote{\url{http://www.vision.caltech.edu/Image_Datasets/Caltech101/}}, \textit{AwA}\footnote{\url{https://cvml.ist.ac.at/AwA/}},
\textit{NUS-Wide}\footnote{\url{https://lms.comp.nus.edu.sg/wp-content/uploads/2019/research/nuswide/NUS-WIDE.html}},
\textit{SunRGBD}\footnote{\url{http://rgbd.cs.princeton.edu/}} and 
\textit{YoutubeFace}\footnote{\url{http://archive.ics.uci.edu/ml/datasets/YouTube+Multiview+Video+Games+Dataset}} (shorted as \textit{YtFace}), which cover a wide range of scales and data types. Together with our newly-built multi-view dataset, we believe that we are able to provide a comprehensive evaluation platform for multi-view deep OCC.  To facilitate future research, we process all multi-view datasets into a consistent format and provide them on a website\footnote{\url{ https://github.com/liujiyuan13/MvDOCC-datasets}}. A summary of all multi-view benchmark datasets used in this paper is given in supplementary material.

\begin{table*}[t]
\centering
\caption{AUROC ($\%$) of different baselines on image based multi-view datasets (best performer in boldface).}  
\begin{tabular}{llcccccccc} 
\toprule
\multicolumn{2}{l}{Type} & MNIST & FashionMNIST & Cifar10 & Cifar100 & SVHN & Cat\_vs\_Dog & MedMNIST & MvTecAD \\ \midrule
\multirow{4}{*}{Fusion} & SUM & 97.57 & 92.76 & 81.10 & \textbf{76.04} & 77.89 & 97.88 & 78.63 & \textbf{89.83} \\
 & MAX & 97.53 & 92.60 & 80.24 & 75.41 & 77.37 & 97.89 & 78.62 & 89.62 \\
 & NN & 97.61 & 92.87 & 79.61 & 75.42 & 77.93 & 97.99 & 78.68 & 89.22 \\
 & TF & 97.58 & 92.54 & 80.60 & 75.47 & 78.40 & 97.74 & 78.42 & 89.06 \\ \midrule
\multirow{3}{*}{Alignment} & DIS & 97.52 & 92.69 & 80.77 & 75.72 & 78.80 & 97.96 & 78.86 & 89.28 \\
 & SIM & 97.57 & 92.67 & 81.22 & 76.00 & 79.03 & 98.01 & 78.71 & 89.58 \\
 & DCCA & 97.59 & 91.97 & 76.53 & 73.58 & \textbf{79.06} & 95.11 & 78.78 & 89.81 \\ \midrule
\multirow{2}{*}{Tailored} & DAE & 97.57 & 92.64 & 81.05 & 75.70 & \textbf{79.06} & \textbf{98.02} & 78.52 & 89.44 \\
 & DSV & 97.20 & 91.89 & 75.15 & 70.70 & 70.48 & 86.07 & 77.57 & 80.43 \\ \midrule
\multirow{2}{*}{Self-supervision} & PPRD & 97.51 & 92.76 & 80.31 & 74.89 & 75.93 & 97.72 & 79.35 & 89.53 \\ 
 & SPRD & \textbf{97.63} & \textbf{92.98} & \textbf{81.30} & 75.52 & 77.29 & 97.90 & \textbf{79.59} & 89.50 \\ \bottomrule
\end{tabular}
\label{tab:auroc_img}
\end{table*}

\begin{table*}[]
\centering
\caption{AUROC ($\%$)  of different baselines on video based multi-view datasets (best performer in boldface).}  
\begin{tabular}{llccccccc} \toprule
\multicolumn{2}{l}{Type} & UCSDped1 & UCSDped2 & UMN\_scene1 & UMN\_scene2 & UMN\_scene3 & Avenue & ShanghaiTech \\ \midrule
\multirow{4}{*}{Fusion} & SUM & \textbf{83.26} & 86.66 & 97.99 & 88.08 & 90.59 & 84.26 & \textbf{67.41} \\
 & MAX & 81.48 & 83.85 & 97.70 & 87.08 & 89.73 & 83.54 & 64.75 \\
 & NN & 82.19 & 83.81 & 98.15 & 87.35 & 90.55 & 83.92 & 67.13 \\
 & TF & 82.95 & 86.17 & 98.12 & 87.78 & 90.82 & \textbf{84.28} & 66.86 \\ \midrule
\multirow{3}{*}{Alignment} & DIS & 81.08 & 82.18 & 97.28 & 86.89 & 90.35 & 82.53 & 64.35 \\
 & SIM & 82.92 & 84.12 & 97.32 & 86.74 & 89.35 & 79.08 & 65.53 \\
 & DCCA & 77.61 & 84.62 & 96.79 & 86.18 & 89.68 & 82.56 & 65.77 \\ \midrule
\multirow{2}{*}{Tailored} & DAE & 80.51 & 81.86 & 97.01 & 87.47 & 88.91 & 83.35 & 64.93 \\
 & DSV & 66.30 & 87.02 & 98.47 & 83.49 & 94.03 & 83.35 & 59.12 \\ \midrule
\multirow{2}{*}{Self-supervision} & PPRD & 78.40 & \textbf{89.62} & 98.45 & 86.31 & \textbf{94.76} & 80.73 & 49.89 \\
 & SPRD & 79.14 & 88.49 & \textbf{98.51} & \textbf{88.41} & 93.12 & 82.11 & 56.09 \\ \bottomrule
\end{tabular}
\label{tab:auroc_video}
\end{table*}

\section{Empirical Evaluations}

Having established formulation, baselines and benchmark datasets for multi-view deep OCC, we perform empirical evaluations to give the first glimpse into this new topic. In addition to head-to-head performance comparison between different baselines, we also conduct in-depth analysis on the characteristics of each model.

\subsection{Experimental Setup}

For multi-view datasets that are specifically designed for OCC (\textit{MvTecAD} and video based multi-view datasets), we directly use the given positive class to train the OCC model, and data from the negative class are used to evaluate OCC performance. For other binary or multi-class multi-view datasets, we apply the ``one \textit{v.s.} all'' protocol (detailed in Sec. \ref{sec:other_datasets}) for training and evaluating the OCC performance. For multi-class datasets that possess more than 10 classes, we select the first 10 qualified classes ($\geq 300$ training data) for experiments. For those multi-view datasets that have already provided the train/test split, we simply use the data of positive class in the training set to train the OCC model, and the test set is used to evaluate the OCC performance. As to those datasets that do not provide train/test split, we randomly sample $70\%$ data of the current positive class as the training set, while the rest of positive class data are mixed with data of the negative class to serve as the testing set. The sampling process is repeated for ten times and the average performance is reported. Before training, training data from each view are normalized to the interval $[-1, 1]$, while the testing set is similarly normalized by the statistics (i.e. min-max value) of the training set. For inference, the reconstruction error based scores of each view are further normalized by the input data dimension, which aims to make scores from different views share the same scale, so they can be comparable and applicable to averaging based late fusion. To quantify the OCC performance, we follow the deep OCC literature and utilize three commonly-used threshold-independent metrics: Area under the Receiver Operation Characteristic Curve (AUROC), Area under the Precision-Recall Curve (AUPR) and True Negative Rate at $95\%$ True Positive Rate (TNR@$95\%$TPR). We also provide more implementation details in the supplementary material, and all of our implementations can be found at \url{ https://github.com/liujiyuan13/MvDOCC-code}.

\subsection{Head-to-head Comparison of Baselines}
\label{sec:comp_baselines}
We test the designed 11 multi-view deep OCC solutions on both our new multi-view datasets and selected existing multi-view datasets. Due to the page limit, we report the most frequently-used AUROC of each baseline for the head-to-head comparison, while the results under other metrics are provided in the supplementary material. Since other metrics actually exhibit a similar trend to AUROC, we will focus on discussing the AUROC performance in this section. The experimental results on image based multi-view datasets, video based multi-view datasets and selected existing multi-view datasets are given in Table \ref{tab:auroc_img}-\ref{tab:auroc_existing}. Note that the performance of \textit{MedMNIST} and \textit{MvTecAD} is given by averaging the performance of each datasets in a collection (detailed results of each dataset in those dataset collections is reported in supplementary material). From those results, we can draw the following observations:

\begin{table*}[t]
\centering
\caption{AUROC ($\%$) of different baselines on selected existing multi-view datasets. The value in the bracket is the $p$-value of student-$t$ test, while $p<0.05$ indicates a significant difference from the best performer.}  
\begin{tabular}{lccccccccccc} \toprule
\multicolumn{1}{l}{} & BBC & BDGP & Caltech20 & Citeseer & Cora & Reuters & Wiki & AwA & NUS-Wide & SunRGBD & YtFace \\ \midrule
SUM & $\text{94.35}_{\pm\text{0.54}}$ & $\text{81.27}_{\pm\text{0.77}}$ & $\text{99.76}_{\pm\text{0.11}}$ & $\text{83.85}_{\pm\text{0.33}}$ & $\text{87.79}_{\pm\text{0.53}}$ & $\text{65.05}_{\pm\text{0.43}}$ & $\text{88.84}_{\pm\text{0.80}}$ & $\text{63.15}_{\pm\text{0.72}}$ & $\text{67.94}_{\pm\text{0.54}}$ & $\text{84.81}_{\pm\text{0.45}}$ & $\text{88.12}$ \\
  & (1.00) & (0.06) & (0.23) & (0.89) & (0.98) & (0.88) & (0.00) & (0.32) & (0.02) & (1.00) & (-) \\
 MAX & $\text{94.35}_{\pm\text{0.54}}$ & $\text{81.36}_{\pm\text{0.84}}$ & $\text{99.69}_{\pm\text{0.17}}$ & $\text{83.86}_{\pm\text{0.33}}$ & $\text{87.79}_{\pm\text{0.51}}$ & $\text{65.04}_{\pm\text{0.42}}$ & $\text{88.93}_{\pm\text{0.64}}$ & $\text{63.34}_{\pm\text{0.77}}$ & $\text{68.25}_{\pm\text{0.49}}$ & $\text{84.63}_{\pm\text{0.51}}$ & $\text{87.86}$ \\
  & (1.00) & (0.10) & (0.05) & (0.96) & (0.98) & (0.85) & (0.00) & (0.64) & (0.15) & (0.60) & (-) \\
 NN & $\text{94.35}_{\pm\text{0.54}}$ & $\text{80.98}_{\pm\text{0.84}}$ & $\text{99.77}_{\pm\text{0.11}}$ & $\text{83.87}_{\pm\text{0.34}}$ & $\text{87.78}_{\pm\text{0.52}}$ & $\text{65.03}_{\pm\text{0.42}}$ & $\text{88.85}_{\pm\text{0.51}}$ & $\text{63.27}_{\pm\text{0.71}}$ & $\text{68.56}_{\pm\text{0.47}}$ & $\text{84.55}_{\pm\text{0.42}}$ & $\text{87.98}$ \\
  & (1.00) & (0.01) & (0.27) & (0.99) & (0.97) & (0.81) & (0.00) & (0.50) & (0.81) & (0.42) & (-) \\
 TF & $\text{94.35}_{\pm\text{0.54}}$ & $\text{81.03}_{\pm\text{0.86}}$ & $\text{97.79}_{\pm\text{1.30}}$ & $\text{83.87}_{\pm\text{0.32}}$ & $\text{87.78}_{\pm\text{0.52}}$ & $\text{65.05}_{\pm\text{0.42}}$ & $\text{89.24}_{\pm\text{1.03}}$ & $\text{62.76}_{\pm\text{0.72}}$ & $\text{67.24}_{\pm\text{0.56}}$ & $\text{84.37}_{\pm\text{0.56}}$ & $\text{86.41}$ \\
  & (1.00) & (0.02) & (0.00) & (0.98) & (0.96) & (0.88) & (0.00) & (0.05) & (0.00) & (0.25) & (-) \\ \midrule
 DIS & $\text{94.35}_{\pm\text{0.54}}$ & $\text{82.03}_{\pm\text{0.80}}$ & $\text{99.82}_{\pm\text{0.08}}$ & $\text{83.86}_{\pm\text{0.34}}$ & $\text{87.79}_{\pm\text{0.53}}$ & $\text{65.05}_{\pm\text{0.42}}$ & $\text{86.58}_{\pm\text{0.77}}$ & $\text{62.96}_{\pm\text{0.67}}$ & $\text{66.91}_{\pm\text{0.64}}$ & $\text{84.16}_{\pm\text{0.43}}$ & $\text{88.41}$ \\
 & (1.00) & (1.00) & (1.00) & (0.94) & (0.96) & (0.90) & (0.00) & (0.13) & (0.00) & (0.07) & (-) \\
 SIM & $\text{94.35}_{\pm\text{0.54}}$ & $\text{81.85}_{\pm\text{0.80}}$ & $\text{99.77}_{\pm\text{0.12}}$ & $\text{83.87}_{\pm\text{0.32}}$ & $\text{87.78}_{\pm\text{0.53}}$ & $\text{65.08}_{\pm\text{0.42}}$ & $\text{86.11}_{\pm\text{0.77}}$ & $\text{62.67}_{\pm\text{0.65}}$ & $\text{67.02}_{\pm\text{0.42}}$ & $\text{84.27}_{\pm\text{0.59}}$ & $\text{88.42}$ \\
  & (1.00)& (0.64) & (0.27) & (0.98) & (0.97) & (1.00) & (0.00) & (0.02) & (0.00) & (0.11) & (-) \\
 DCCA & $\text{94.35}_{\pm\text{0.54}}$ & $\text{81.74}_{\pm\text{0.84}}$ & $\text{99.74}_{\pm\text{0.14}}$ & $\text{83.86}_{\pm\text{0.34}}$ & $\text{87.78}_{\pm\text{0.52}}$ & $\text{65.08}_{\pm\text{0.42}}$ & $\text{87.49}_{\pm\text{0.84}}$ & $\text{62.76}_{\pm\text{0.67}}$ & $\text{66.89}_{\pm\text{0.48}}$ & $\text{84.00}_{\pm\text{0.49}}$ & $\text{87.72}$ \\
  & (1.00) & (0.47) & (0.15) & (0.97) & (0.97) & (1.00) & (0.00) & (0.04) & (0.00) & 0.04 & (-) \\ \midrule
 DAE & $\text{94.35}_{\pm\text{0.54}}$ & $\text{81.99}_{\pm\text{0.79}}$ & $\text{99.80}_{\pm\text{0.11}}$ & $\text{83.86}_{\pm\text{0.32}}$ & $\text{87.79}_{\pm\text{0.52}}$ & $\text{65.05}_{\pm\text{0.42}}$ & $\text{85.87}_{\pm\text{0.56}}$ & $\text{62.84}_{\pm\text{0.70}}$ & $\text{66.59}_{\pm\text{0.57}}$ & $\text{84.18}_{\pm\text{0.43}}$ & $\text{88.33}$ \\
  & (1.00) & (0.93) & (0.60) & (0.95) & (1.00) & (0.87) & (0.00) & (0.07) & (0.00) & (0.08) & (-) \\
 DSV & $\text{93.64}_{\pm\text{0.59}}$ & $\text{76.09}_{\pm\text{1.51}}$ & $\text{98.11}_{\pm\text{0.24}}$ & $\text{72.86}_{\pm\text{0.65}}$ & $\text{82.66}_{\pm\text{1.08}}$ & $\text{64.53}_{\pm\text{0.40}}$ & $\text{84.81}_{\pm\text{0.54}}$ & $\text{61.96}_{\pm\text{0.47}}$ & $\text{66.33}_{\pm\text{0.78}}$ & $\text{68.33}_{\pm\text{1.22}}$ & $\text{90.04}$ \\
 & (0.02) & (0.00) & (0.00) & (0.00) & (0.01) & (0.00) & (0.00) & (0.00) & (0.00) & (0.00) & (-) \\ \midrule
 PPRD & $\text{94.35}_{\pm\text{0.54}}$ & $\text{81.13}_{\pm\text{1.00}}$ & $\text{99.55}_{\pm\text{0.20}}$ & $\text{83.86}_{\pm\text{0.33}}$ & $\text{87.78}_{\pm\text{0.51}}$ & $\text{65.03}_{\pm\text{0.42}}$ & $\text{90.93}_{\pm\text{0.53}}$ & $\text{63.51}_{\pm\text{0.79}}$ & $\text{67.71}_{\pm\text{0.40}}$ & $\text{83.39}_{\pm\text{0.40}}$ & $\text{88.25}$ \\
  & (1.00) & (0.05) & (0.00) & (0.94) & (0.96) & (0.79) & (1.00) & (1.00) & (0.00) & (0.00) & (-) \\
 SPRD & $\text{94.35}_{\pm\text{0.54}}$ & $\text{79.50}_{\pm\text{0.93}}$ & $\text{99.61}_{\pm\text{0.19}}$ & $\text{83.87}_{\pm\text{0.33}}$ & $\text{87.78}_{\pm\text{0.52}}$ & $\text{65.01}_{\pm\text{0.42}}$ & $\text{90.82}_{\pm\text{0.63}}$ & $\text{63.50}_{\pm\text{0.64}}$ & $\text{68.62}_{\pm\text{0.55}}$ & $\text{84.81}_{\pm\text{0.44}}$ & $\text{87.67}$ \\
  & (1.00) & (0.00) & (0.01) & (1.00) & (0.95) & (0.75) & (0.70) & (0.98) & (1.00) & (1.00) & (-) \\ \bottomrule
\end{tabular}
\label{tab:auroc_existing}
\end{table*}

1) \textit{In some cases, most of baseline solutions actually achieve fairly close performance}, despite of their difference in type and implementation. Concretely, as shown in Table \ref{tab:auroc_existing}, baseline solutions attain almost identical performance on several existing multi-view datasets that are widely-used in the literature, e.g. \textit{BBC}, \textit{Caltech20}, \textit{Cora} and \textit{Reuters}. On many image based multi-view datasets, we also note that the best performer usually leads other counterparts by a less than $1\%$ AUROC. However, baselines could also obtain evidently different performance on other multi-view datasets, like some video based and image based datasets. This also justifies the necessity for a comprehensive evaluation. 2) \textit{There does not exist a single baseline that can consistently outperform other baselines}. For example, we notice that self-supervision based baselines (PPRD and SPRD) attains the optimal or near-optimal performance (i.e. not significantly different from the best performer) on $16$ out of the total $26$ datasets (\textit{MedMNIST} and \textit{MvTecAD} are viewed as two datasets here). However, self-supervision based baselines also suffer from evidently inferior performance to other baselines on some datasets, such as \textit{UCSDped1} and \textit{ShanghaiTech}. 3) \textit{Simple fusion functions (SUM and MAX) can readily compete with comparatively complex fusion functions (NN and TF)}. In fact, all fusion based baselines yield fairly comparable performance on most datasets. To our surprise, summation turns out to be the most effective way to conduct fusion in our evaluation. 4) \textit{Correlation based alignment undergoes more fluctuations than other ways of alignment}. It can be observed that DCCA based alignment sometimes performs evidently worse than its two alignment based counterparts, e.g. on \textit{Cifar10/Cifar100}, \textit{Cat\_vs\_Dog} and \textit{YtFace}. By contrast, distance based alignment maintains the most stable performance in the evaluation. 5) \textit{DAE proves to be a strong baseline, while the performance of DSV is typically unsatisfactory in most cases}. Although DAE is a simple extension from the single-view deep autoencoder, it is able to produce acceptable or even superior performance to other baselines that are more sophisticated. However, DSVDD based baseline often achieves lower AUROC than other baselines, although it is the best performer on the recent \textit{YtFace} dataset.

In the supplementary material, we also show the performance of baselines under other metrics (AUPR, TNR@$95\%$TPR), as well as the performance of four miscellaneous baselines. We believe those results lay a firm foundation for future research on multi-view deep OCC.




\begin{figure*}[ht]
\centering
\includegraphics[width=\linewidth]{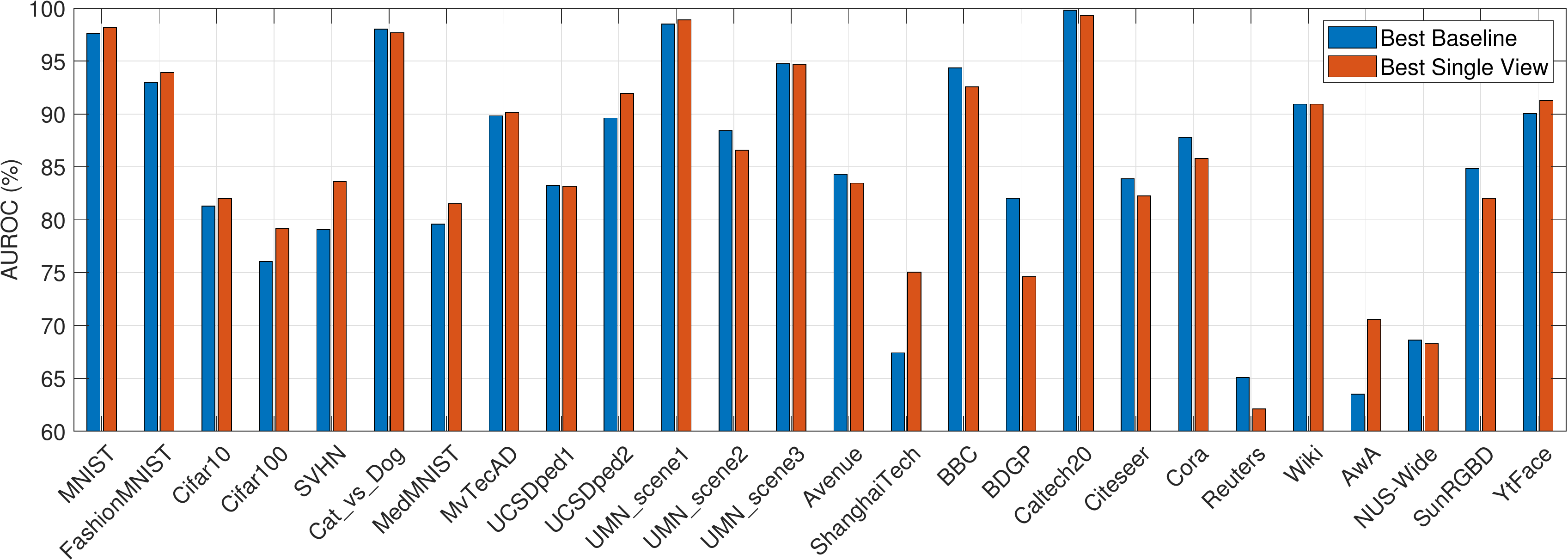}
\caption{AUROC ($\%$) between the best baseline and best single view.}
\label{fig:mvdooc_single_best}
\end{figure*}

\begin{figure*}[ht]
\centering
\includegraphics[width=\linewidth]{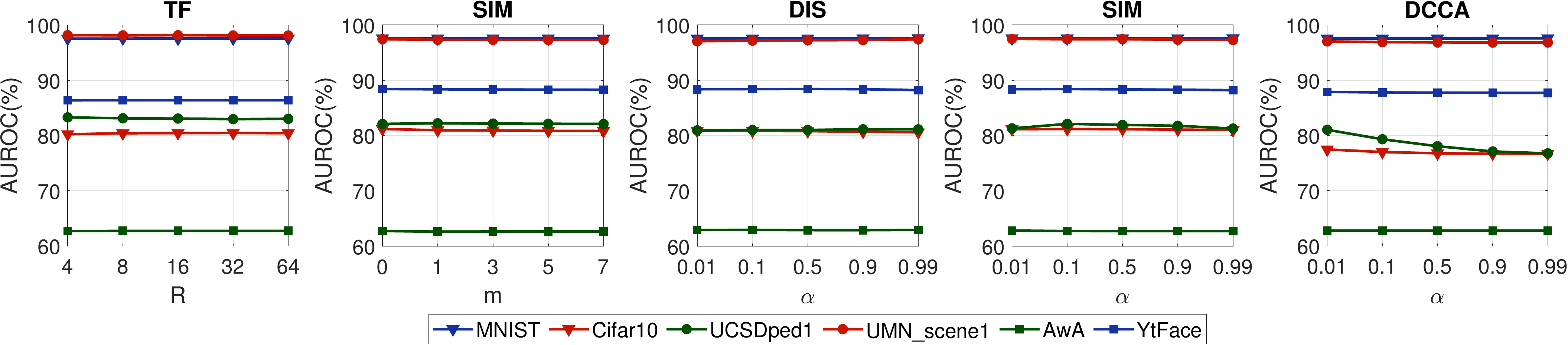}
\caption{Sensitivity analysis of typical hyperparameters in multi-view deep OCC.}
\label{fig:param}
\end{figure*}

\begin{table*}[t]
\centering
\caption{AUROC ($\%$) of different late fusion strategies on selected existing multi-view datasets.} 
\label{tab:roc_late_fusion}
\begin{tabular}{lcccccccccc} \toprule
 & BBC & BDGP & Caltech20 & Citeseer & Cora & Reuters & Wiki & AwA & NUS-Wide & SunRGBD \\ \midrule
LF-AVG & \textbf{$\text{94.35}_{\pm\text{0.54}}$} & $\text{81.99}_{\pm\text{0.79}}$ & \textbf{$\text{99.80}_{\pm\text{0.11}}$} & \textbf{$\text{83.86}_{\pm\text{0.32}}$} & \textbf{$\text{87.79}_{\pm\text{0.52}}$} & \textbf{$\text{65.05}_{\pm\text{0.42}}$} & $\text{85.87}_{\pm\text{0.56}}$ & \textbf{$\text{62.84}_{\pm\text{0.70}}$} & \textbf{$\text{66.59}_{\pm\text{0.57}}$} & \textbf{$\text{84.18}_{\pm\text{0.43}}$} \\
LF-MIN & $\text{93.24}_{\pm\text{0.64}}$ & \textbf{$\text{82.70}_{\pm\text{0.87}}$} & $\text{99.44}_{\pm\text{0.20}}$ & $\text{81.99}_{\pm\text{0.34}}$ & $\text{82.35}_{\pm\text{0.74}}$ & $\text{63.36}_{\pm\text{0.39}}$ & $\text{79.94}_{\pm\text{0.72}}$ & $\text{61.80}_{\pm\text{0.70}}$ & $\text{63.88}_{\pm\text{0.47}}$ & $\text{81.48}_{\pm\text{0.67}}$ \\
LF-MAX & $\text{94.11}_{\pm\text{0.41}}$ & $\text{51.64}_{\pm\text{1.24}}$ & $\text{91.91}_{\pm\text{0.99}}$ & $\text{53.11}_{\pm\text{0.76}}$ & $\text{55.88}_{\pm\text{0.51}}$ & $\text{59.74}_{\pm\text{0.22}}$ & \textbf{$\text{90.49}_{\pm\text{0.84}}$} & $\text{54.93}_{\pm\text{0.59}}$ & $\text{64.44}_{\pm\text{0.50}}$ & $\text{82.16}_{\pm\text{0.36}}$ \\ \bottomrule
\end{tabular}
\end{table*}

\subsection{Further Analysis}

\subsubsection{Comparison with Single-view Performance}
\label{sec:best_single}
To enable a better insight to devised multi-view deep OCC baselines, we conduct an experiment to compare best baselines' performance and the best single-view performance on each benchmark dataset in terms of AUROC. The best single-view performance is obtained by training a deep autoencoder with data from one single views and selecting the best performer among the obtained deep autoencoders. In particular, it should be noted that the best single-view performance is actually hindsight, i.e. it is usually not practically accessible due to the absence of the negative class in multi-view deep OCC. Therfore, it is merely used as a reference to reflect how existing baselines exploit multi-view information. The results are shown in Fig. \ref{fig:mvdooc_single_best}, and we can come to an interesting conclusion: \textit{Despite of a systematic exploration, current baselines still suffer from insufficient capability to exploit multi-view information for multi-view deep OCC.} Specifically, on 12 out of the total 26 datasets, the performance of best baseline is still inferior to the best single-view performance. However, an ideal multi-view learning model is supposed to be superior or comparable to the best single-view performance. Such results imply two facts: First, it is discovered that existing baselines are still unable to find a perfect way to exploit the contributing information embedded in each view. Second, redundant information in multi-view data could be detrimental to the multi-view deep OCC performance. As a consequence, there is a large room for developing improved multi-view deep OCC solutions.

\subsubsection{Sensitivity Analysis}

In this section, we will discuss the impact of typical hyperparameters for the devised baselines: The rank number $R$ for tensor based fusion, the margin $m$ for similarity based fusion and the weight of alignment loss $\alpha$ for alignment based baselines (DIS, SIM and DCCA). We choose the $R$, $m$ and $\alpha$ value from $\{4, 8, 16, 32, 64\}$, $\{0, 1, 3, 5, 7\}$ and $\{0.01, 0.1, 0.5, 0.9, 0.99\}$ respectively, and show the corresponding performance on representative datasets in Fig. \ref{fig:param}. Surprisingly, we notice that the performance under different hyperparameter settings remains stable in the majority of cases. The performance fluctuations are usually within the range of $1\%$, except for the case of DCCA on UCSDped1. Consequently, we can speculate that a breakthrough of performance requires progress on model design, and tuning hyperparameter may not produce a performance leap.

\subsubsection{Influence of Late Fusion}

As a common component for almost all baselines, late fusion has a major influence on the performance multi-view deep OCC. Since we assume that no datum from negative classes are available for validation, it is hard to apply many existing late fusion solutions here. As a preliminary effort, we take DAE for an example and explore three simple strategies for late fusion: Averaging strategy (LF-AVG, used by default), max-value strategy (LF-MAX) and min-value strategy (LF-MIN), which compute the final score by the mean, maximum and minimum of all views' scores. For simplicity, we test them on existing multi-view datasets and show the results in Table \ref{tab:roc_late_fusion}. As it is shown by Table \ref{tab:roc_late_fusion}, the averaging strategy almost constantly outperforms max-value and min-value strategy (except for \textit{BDGP} and \textit{Wiki}). The min-value strategy also achieves acceptable results in most cases, which is consistent with our intuition that any abnormal view should signify an abnormal datum. However, it is noted that the max-value strategy can produce very poor fusion results, e.g. on \textit{Citeseer} and \textit{Cora} dataset. Therefore, the averaging strategy could still be an informative baseline late fusion strategy for multi-view deep OCC, which is somewhat similar to the case of multi-view learning.

\section{Discussion}

Based on the results of previous experiments, we would like to make the following remarks on the multi-view deep OCC, which may inspire further research on this new topic:

\begin{itemize}
    \item \textit{A non-trivial ``killer'' approach to multi-view deep OCC still requires exploration.} As we have shown in Sec. \ref{sec:comp_baselines}, there is not a single baseline that can consistently outperforms its counterparts. In the meantime, the performance gap between different baselines can be very small in many cases. Thus, it will be very attractive to explore the possibility to design a new multi-view deep OCC solution. In particular, we believe that self-supervised learning can be a promising direction to find such a solution, considering its comparatively better performance among baselines and the remarkable progress achieved by the self-supervised learning community.
    
    \item \textit{It will be interesting to assess the quality or contribution of each view to multi-view deep OCC.} Since prior knowledge on negative classes is not given, it will be natural to describe a sample by as many views as possible. However, as it is shown in Sec. \ref{sec:best_single}, it may degrade the performance when data of multiple views are blindly fused or aligned. Therefore, it is of high value to develop a strategy to perform knowledgeable multi-view fusion or alignment. This is also applicable to the late fusion stage.
    
    \item \textit{The revolution of the learning paradigm may breed a breakthrough.} The generative learning paradigm (i.e. generation or prediction) has been a standard practice in deep OCC, which is followed in this paper when designing most baselines. However, other learning paradigms, such as the discriminative learning \cite{golan2018deep} and contrastive learning \cite{Xiao2020Self} paradigm, have proved to be more effective than generative learning paradigm in realms like unsupervised representation learning. Naturally, a brand-new learning paradigm may be a good remedy to multi-view deep OCC.
    
    \item \textit{Newly-emerging DNN models can be explored for enhancing multi-view deep OCC.} In this paper, most baselines are developed based on the classic encoder-decoder like DNN models. This is due to the fact that deep autoencoder and its variants are the most commonly-used tool for deep OCC, and they can be good reference to understand multi-view deep OCC. However, the deep OCC realm also witness the emergence of many emerging DNN models, such as GANs \cite{gui2020review} and transformers \cite{2020ViT}. Such new techniques pave the way for better multi-view deep OCC. For example, it will be interesting to leverage the self-attention mechanism of transformers to capture the inter-view correspondence within multi-view data.
\end{itemize}


\section{Conclusion}
\label{sec:conclusion}

This paper investigates a pervasive but unexplored problem: Multi-view deep OCC. Within the scope of our best knowledge, we are the first to formally identify and formulate multi-view deep OCC. In order to overcome the practical difficulties to look into this problem, we systematically design baseline solutions by extensively reviewing relevant areas in the literature, and we also construct abundant new multi-view datasets by processing public data via various means. Together with some existing multi-view datasets, a comprehensive evaluation of designed baselines is carried out to provide the first glimpse to this new topic. To facilitate future research on this challenging problem, all benchmark datasets and implementations of baselines are open.  


\bibliographystyle{IEEEtran}
\bibliography{mvdocc}

\clearpage

\end{document}